\documentclass[11pt,a4paper]{article}
\usepackage[hyperref]{acl2021}
\usepackage{times}
\usepackage{latexsym}

\usepackage{multicol}
\usepackage{multirow}
\usepackage{booktabs}
\usepackage{makecell}
\usepackage{float}
\usepackage{enumitem}
\usepackage{graphicx}
\usepackage{amsmath,amssymb}
\usepackage{xcolor}
\usepackage{hyperref}
\usepackage{todonotes}
\usepackage{diagbox}
\usepackage{tabu}
\usepackage{url}
\usepackage[export]{adjustbox}
\usepackage{subcaption}
\usepackage{xspace}
\usepackage{colortbl}
\usepackage{stfloats}

\definecolor{customgreen}{RGB}{70,154,37}

\newcommand{\model}{{\textsc{TCS}}\xspace}
\newcommand{\modelsupervised}{{\textsc{TCS (S)}}\xspace}
\newcommand{\modelunsupervised}{{\textsc{TCS (U)}}\xspace}
\newcommand{\modellex}{{\textsc{TCS (LEX)}}\xspace}
\newcommand{\modelsimalign}{{\textsc{TCS (simalign)}}\xspace}
\newcommand{\combinedcs}{{{All-CS}}\xspace}
\newcommand{\moviecs}{{{Movie-CS}}\xspace}
\newcommand{\treebankcs}{{{Treebank-CS}}\xspace}
\newcommand{\opus}{{{OpSub}}\xspace}
\newcommand{\opushundred}{{{OPUS-100}}\xspace}
\newcommand{\masktodo}[1]{}
\newcommand*\samethanks[1][\value{footnote}]{\footnotemark[#1]}
\usepackage{microtype}
\usepackage{microtype}

\aclfinalcopy % Uncomment this line for the final submission
\def\aclpaperid{***} %  Enter the acl Paper ID here

\title{From Machine Translation to Code-Switching:\\ Generating High-Quality Code-Switched Text}

\author{ Ishan Tarunesh\thanks{ \hspace{5pt}Work done while first two authors were students at IIT Bombay.}, Syamantak Kumar\samethanks, Preethi Jyothi \\
  Samsung Korea, Google India, IIT Bombay\\
  \texttt{\{ishantarunesh, syamantak.kumar\}@gmail.com, pjyothi@cse.iitb.ac.in}}

\date{}

\begin{document}

\maketitle
\begin{abstract}
Generating code-switched text is a problem of growing interest, especially given the scarcity of corpora containing large volumes of real code-switched text. In this work, we adapt a state-of-the-art neural machine translation model to generate Hindi-English code-switched sentences starting from monolingual Hindi sentences. We outline a carefully designed curriculum of pretraining steps, including the use of synthetic code-switched text, that enable the model to generate high-quality code-switched text. Using text generated from our model as data augmentation, we show significant reductions in perplexity on a language modeling task, compared to using text from other generative models of CS text. We also show improvements using our text for a downstream code-switched natural language inference task. Our generated text is further subjected to a rigorous evaluation using a human evaluation study and a range of objective metrics, where we show performance comparable (and sometimes even superior) to code-switched text obtained via crowd workers who are native Hindi speakers. 
\end{abstract}

\section{Introduction}
\label{section:intro}

Code-switching (CS) refers to the linguistic phenomenon of using more than one language within a single sentence or conversation. CS appears naturally in conversational speech among multilingual speakers. %With the rise in technologies like voice-assisted interfaces and chat bots, building computational models for conversational CS text has emerged as an important problem. %it also appears in textual form on social media platforms like Twitter and Facebook.
The main challenge with building models for conversational CS text is that we do not have access to large amounts of CS text that is conversational in style. One might consider using social media text that contains CS and is more readily available. However, the latter is quite different from conversational CS text in its vocabulary (e.g., due to the frequent use of abbreviated slang terms, hashtags and mentions), in its sentence structure (e.g., due to character limits in tweets) and in its word forms (e.g., due to transliteration being commonly employed in social media posts). This motivates the need for a generative model of realistic CS text that can be sampled to subsequently train models for CS text.

In this work, we tackle the problem of generating high-quality CS text using only limited amounts of real CS text during training. We also assume access to large amounts of monolingual text in the component languages and parallel text in both languages, which is a reasonable assumption to make for many of the world's languages. We focus on Hindi-English CS text where the matrix (dominant) language is Hindi and the embedded language is English.%
\footnote{Given the non-trivial effort involved in collecting annotations from professional annotators and crowd workers, we focused on a single language pair (Hindi-English) and leave explorations on more language pairs for future work.}
Rather than train a generative model, we treat this problem as a translation task where the source and target languages are monolingual Hindi text and Hindi-English CS text, respectively. We also use the monolingual Hindi text to construct synthetic CS sentences using simple techniques. We show that synthetic CS text, albeit being naive in its construction, plays an important role in improving our model's ability to capture CS patterns. %We empirically demonstrate that our model is able to produce sentences of much higher quality compared to prior work involving generative models.

We draw inspiration from the large body of recent work on unsupervised machine translation~\cite{unmt,lample-etal-2018-phrase} to design our model, which will henceforth be referred to as \textbf{T}ranslation for \textbf{C}ode-\textbf{S}witching, or \model. \model, once trained, will convert a monolingual Hindi sentence into a Hindi-English CS sentence. \model makes effective use of parallel text when it is available and uses backtranslation-based objective functions with monolingual text. 
%We evaluate the quality of text generated via \model using human evaluations and a number of different objective metrics, including training a language model on the generated text and computing perplexities on an unseen test set of CS sentences.

%\model is used to generate CS text, starting from monolingual text, which is subsequently used to train a language model (LM). The trained LM is then used to compute perplexities on an unseen test set of real CS text. We show that sentences generated using \model are much more effective in reducing perplexities compared to other prior approaches. We also conduct human evaluations of the generated CS text, along with defining other objective metrics that measure the quality and diversity of CS patterns in the generated text.  We conduct a thorough evaluation of the quality of the text generated by \model using multiple techniques. 

%To make use of these different parallel, non-parallel and monolingual corpora within a single model, we draw inspiration from the large body of recent work on unsupervised machine translation~\cite{unmt,lample-etal-2018-phrase}. These architectures provide us the means to make effective use of parallel text when its available and use backtranslation with monolingual text. We will henceforth refer to our model as \textbf{T}ranslation for \textbf{C}ode-\textbf{S}witching, or \model. 

Below, we summarize our main contributions:
\begin{enumerate}
\item We propose a state-of-the-art translation model that generates Hindi-English CS text starting from monolingual Hindi text. This model requires very small amounts of real CS text, uses both supervised and unsupervised training objectives and considerably benefits from a carefully designed training curriculum, that includes pretraining with synthetically constructed CS sentences.
\item We introduce a new Hindi-English CS text corpus in this work.%
%\footnote{This new dataset can be downloaded by filling out this \href{https://forms.gle/meTNwNqdrgpjepbUA}{form}. The code is available \href{https://github.com/csalt-research/translation-for-code-switching-acl}{here}}.
\footnote{The new dataset and relevant code is available at:\\ 
\url{https://www.cse.iitb.ac.in/~pjyothi/TCS}.}
Each CS sentence is accompanied by its monolingual Hindi translation. We also designed a crowdsourcing task to collect CS variants of monolingual Hindi sentences. The crowdsourced CS sentences were manually verified and form a part of our new dataset.
\item We use sentences generated from our model to train language models for Hindi-English CS text and show significant improvements in perplexity compared to other approaches.
\item We present a rigorous evaluation of the quality of our generated text using multiple objective metrics and a human evaluation study, and they clearly show that the sentences generated by our model are superior in quality and successfully capture naturally occurring CS patterns.
\end{enumerate}

\section{Related Work}
\label{section:related_word}
%\vspace{-1pt}

Early approaches of language modeling for code-switched text included class-based $n$-gram models~\cite{yeh2010integrated}, factored language models that exploited a large number of syntactic and semantic features~\cite{adel2015syntactic},  and recurrent neural language models~\cite{adel2013recurrent} for CS text. All these approaches relied on access to real CS text to train the language models. Towards alleviating this dependence on real CS text, there has been prior work on learning code-switched language models from bilingual data~\cite{li-fung-2014-language,Li2014CodeSL,garg2018dual} and a more recent direction that explores the possibility of generating synthetic CS sentences. \citep{pratapa2018language} presents a technique to generate synthetic CS text that grammatically adheres to a linguistic theory of code-switching known as the equivalence constraint (EC) theory~\cite{Poplack1979SometimesIS,sankoff_1998}. \newcite{lee-li-2020-modeling} proposed a bilingual attention language model for CS text trained solely using a parallel corpus. 

Another recent line of work has explored neural generative models for CS text. \citet{garg-etal-2018-code} use a sequence generative adversarial network (SeqGAN~\cite{seqgan}) trained on real CS text to generate sentences that are used to aid language model training. Another GAN-based method proposed by~\citet{chang-etal} aims to predict the probability of switching at each token. \citet{learn-to-code-switch} and \citet{pointer-net} use a sequence-to-sequence model enabled with a copy mechanism (Pointer Network~\cite{vinyals2015pointer}) to generate CS data by leveraging parallel monolingual translations from a limited source of CS data. \citet{vacs} proposed a hierarchical variational autoencoder-based model tailored for code-switching that takes into account both syntactic information and language switching signals via the use of language tags. (We present a comparison of \model with both \citet{vacs} and \citet{garg-etal-2018-code} in Section~\ref{section:lm_modelling}.)

In a departure from using generative models for CS text, we view this problem as one of sequence transduction where we train a model to convert a monolingual sentence into its CS counterpart. \citet{chang-etal,gao2019code} use GAN-based models to modify monolingual sentences into CS sentences, while we treat this problem of CS generation as a translation task and draw inspiration from the growing body of recent work on neural unsupervised machine translation models~\cite{unmt,lample-etal-2018-phrase} to build an effective model of CS text. %\citet{chang-etal} required large amounts of CS text to train their GAN-based model. With limited amounts of CS text as in our setup, the outputs would be very poor; hence, we did not compare against it. 

The idea of using translation models for code-switching has been explored in early work~\cite{vu2012first,li2013improved,dhar-etal-2018-enabling}. Concurrent with our work, there have been efforts towards building translation models from English to CS text~\cite{calcs-2021-approaches} and CS text to English~\cite{gupta2021training}. While these works focus on translating from the embedded language (English) to the CS text or vice-versa, our approach starts with sentences in the matrix language (Hindi) which is the more dominant language in the CS text. Also, ours is the first work, to our knowledge, to repurpose an unsupervised neural machine translation model to translate monolingual sentences into CS text. Powerful pretrained models like mBART~\cite{liu2020multilingual} have been used for code-mixed translation tasks in concurrent work~\cite{gautam2021comet}. We will further explore the use of synthetic text with such models as part of future work.
%More importantly, successfully training large neural machine translation models with limited access to real CS text is an important challenge that we tackle in this work.

% \vspace{-1.5pt}

\section{Our Approach}
\label{section:model}

% \vspace{-2pt}

\begin{figure}[t!]
    \centering
    \includegraphics[width=0.5\textwidth]{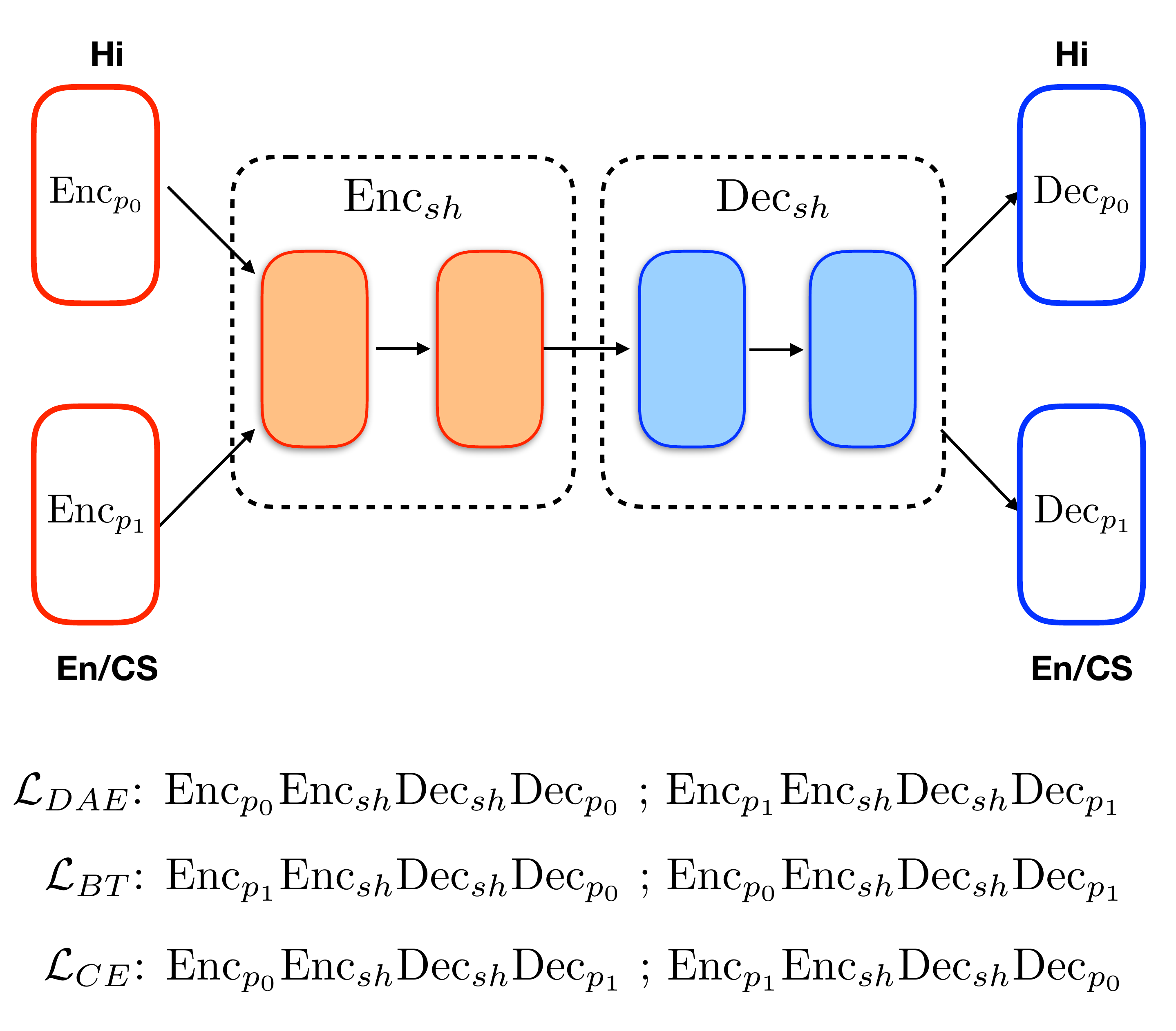}
    \caption{\small Model architecture. Each loss term along with all the network components it modifies are shown. During unsupervised training with non-parallel text, $\mathcal{L}_{DAE}$ and $\mathcal{L}_{BT}$ are optimized while for supervised training with parallel text, $\mathcal{L}_{DAE}$ and $\mathcal{L}_{CE}$ are optimized.}
    \label{fig:my_label}
    %\vspace{-10pt}
\end{figure}

Figure~\ref{fig:my_label} shows the overall architecture of our model. This is largely motivated by prior work on unsupervised neural machine translation~\cite{unmt,lample-etal-2018-phrase}. The model comprises of three layers of stacked Transformer~\cite{transformer} encoder and decoder layers, two of which are shared and the remaining layer is private to each language. Monolingual Hindi (i.e. the source language) has its own private encoder and decoder layers (denoted by $\mathrm{Enc}_{p_0}$ and $\mathrm{Dec}_{p_0}$, respectively) while English and Hindi-English CS text jointly make use of the remaining private encoder and decoder layers (denoted by $\mathrm{Enc}_{p_1}$ and $\mathrm{Dec}_{p_1}$, respectively). In our model, the target language is either English or CS text. Ideally, we would like $\mathrm{Enc}_{p_1}$ and $\mathrm{Dec}_{p_1}$ to be trained only using CS text. However, due to the paucity of CS text, we also use text in the embedded language (i.e. English) to train these layers. Next, we outline the three main training steps of \model.

%Our model is trained using both supervised training objectives (with parallel data) and unsupervised training objectives (with non-parallel data). 
%\vspace{-2pt}
\paragraph{(I) Denoising autoencoding (DAE).} We use monolingual text in each language to estimate language models.  In~\citet{lample-etal-2018-phrase}, this is achieved via denoising autoencoding where an autoencoder is used to reconstruct a sentence given a noisy version as its input whose structure is altered by dropping and swapping words arbitrarily~\cite{unmt}. The loss incurred in this step is denoted by $\mathcal{L}_{DAE}$ and is composed of two terms based on the reconstruction of the source and target language sentences, respectively. 

%\vspace{-2pt}
\paragraph{(II) Backtranslation (BT):} Once the layers are initialized, one can use non-parallel text in both languages to generate a pseudo-parallel corpus of backtranslated pairs~\cite{sennrich}. That is, a corpus of parallel text is constructed by translating sentences in the source language via the pipeline, $\mathrm{Enc}_{p_0}$, $\mathrm{Enc}_{sh}$, $\mathrm{Dec}_{sh}$ and $\mathrm{Dec}_{p_1}$, and translating target sentences back to the source language via $\mathrm{Enc}_{p_1}$, $\mathrm{Enc}_{sh}$, $\mathrm{Dec}_{sh}$ and $\mathrm{Dec}_{p_0}$. The backtranslation loss $\mathcal{L}_{BT}$ is composed of cross-entropy losses from using these pseudo-parallel sentences in both directions.

%\vspace{-2pt}
\paragraph{(III) Cross-entropy loss (CE):} Both the previous steps used unsupervised training objectives and make use of non-parallel text. With access to parallel text, one can use the standard supervised cross-entropy loss (denoted by $\mathcal{L}_{CE}$) to train the translation models (i.e. going from $\mathrm{Enc}_{p_0}$ to $\mathrm{Dec}_{p_1}$ and $\mathrm{Enc}_{p_1}$ to $\mathrm{Dec}_{p_0}$ via the common shared layers).

\subsection{Synthetic CS text} 
\label{ssec:synthCS}

Apart from the use of parallel text and monolingual text employed in training \model, we also construct large volumes of synthetic CS text using two simple techniques. This synthetic CS text is non-parallel and is used to optimize both $\mathcal{L}_{DAE}$ and $\mathcal{L}_{BT}$. The role of the synthetic CS text is to expose \model to various CS patterns (even if noisy), thereby encouraging the model to code-switch. The final step of finetuning using \combinedcs enables model to mimic switching patterns of real CS texts

%We use two simple techniques to create large volumes of synthetic CS text. The idea was to expose \model to various switching patterns in text via the synthetic CS datasets and finally refine the model with smaller amounts of real CS text.  

%Given the lack of real CS text to train the target language-specific modules, we created synthetic CS text using two simple strategies. The resulting text proves to be very useful as a pretraining corpus (as discussed in more detail in our experiments in Section~\ref{section:experiments}).
The first technique (named LEX) is a simple heuristic-based technique that constructs a CS sentence by traversing a Hindi sentence and randomly replacing a word by its English translation using a bilingual lexicon~\cite{conneau2017word}. The probability of replacing a word is chosen to match the switching distribution in real CS text. The second technique (named EMT) is more linguistically aware. Following the methodology proposed by~\citet{Bhat:16}\masktodo{cite Bhat et al. 2016 from MSR} that is based on the embedded matrix theory (EMT) for code-switching, we apply clause substitution methods to monolingual text to construct synthetic CS text. From inspecting English parse trees, we found that replacing embedded sentence clauses or subordinate clauses with their Hindi translations would likely produce CS text that appears somewhat natural.%

\section{Description of Datasets}
\label{section:datasets}

% \vspace{-2pt}
%There are many existing and new datasets used in this work that require a detailed exposition. We will introduce our new Hindi-English CS dataset first, followed by details of datasets curated from existing sources.

\subsection{A New Hindi-English CS Dataset} 
We introduce a new Hindi-English CS dataset, that we will refer to as \combinedcs. %
%\footnote{Interested readers can download All-CS data by filling the \href{https://forms.gle/nvbHCTJPiQVtfGvd8}{following form}.}
%
It is partitioned into two subsets, \moviecs and \treebankcs, based on their respective sources. \moviecs consists of conversational Hindi-English CS text extracted from 30 contemporary Bollywood scripts that were publicly available.%
\footnote{
\href{https://www.filmcompanion.in/category/fc-pro/scripts/}{https://www.filmcompanion.in/category/fc-pro/scripts/}
\href{https://moifightclub.com/category/scripts/}{https://moifightclub.com/category/scripts/}
}
The Hindi words in these sentences were all Romanized with potentially multiple non-canonical forms existing for the same Hindi token. We employed a professional annotation company to convert the Romanized Hindi words into their respective back-transliterated forms rendered in Devanagari script. We also asked the annotators to provide monolingual Hindi translations for all these sentences. Using these monolingual Hindi sentences as a starting point, we additionally crowdsourced for CS sentences via Amazon's Mechanical Turk (MTurk)~\cite{mturk}. %resulting in a total of $19871$ sentences.\todo{Why 19871? Is this correct?}
Table 1 shows two Hindi sentences from \moviecs and \treebankcs, along with the different variants of CS sentences. 
 
Turkers were asked to convert a monolingual Hindi sentence into a natural-sounding CS variant that was semantically identical. Each Turker had to work on five Hindi sentences. We developed a web interface using which Turkers could easily copy parts of the Hindi sentence they wanted to retain and splice in English segments. More details about this interface, the crowdsourcing task and worker statistics are available in Appendix~\ref{sec:appendix_mturk}.

\begin{figure}
    \centering
    \includegraphics[width=0.45\textwidth]{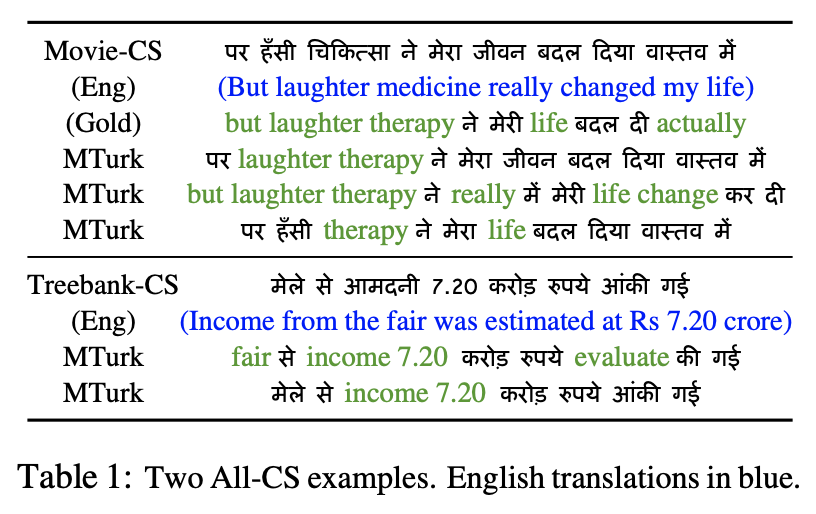}
\end{figure}

\begin{figure}[t!]
  \centering
  \begin{subfigure}[b]{1.1\linewidth}
    \includegraphics[width=0.9\linewidth,height=4cm]{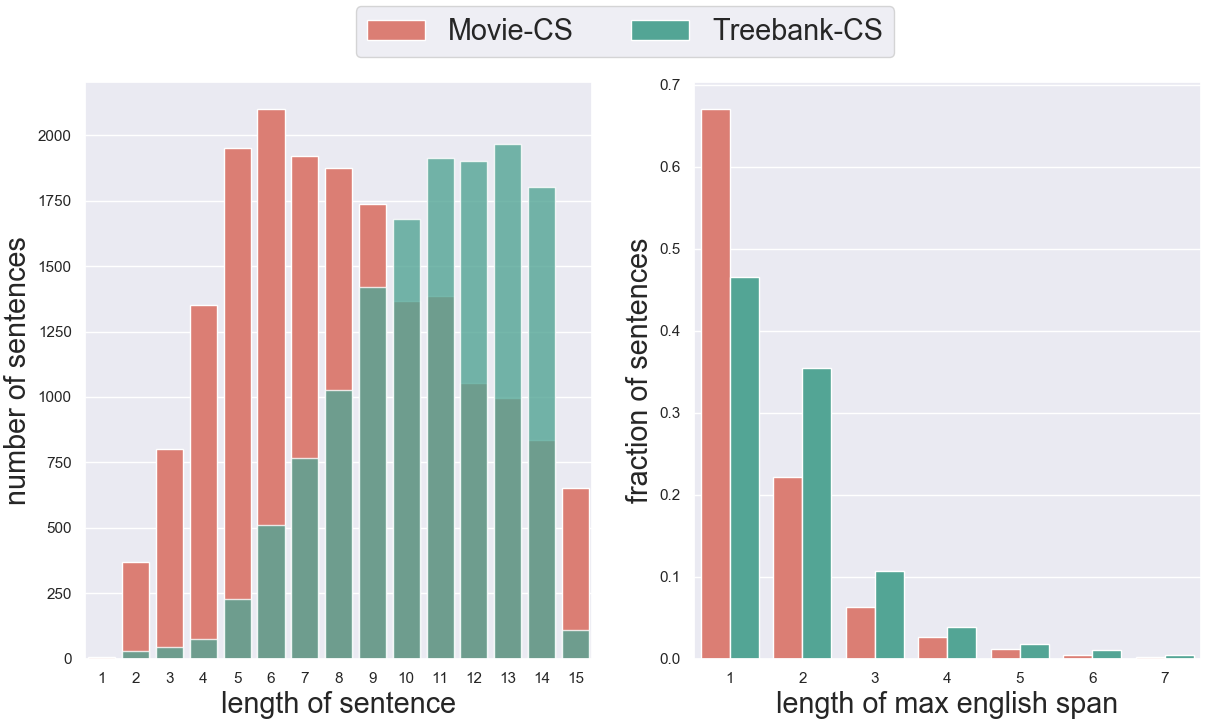}
  \end{subfigure}
  \\
%   \begin{subfigure}[b]{1.1\linewidth}
%     \includegraphics[width=0.9\linewidth]{treebankcs_stats.png}
%      \caption{TreebankCS}
%      \label{fig:fig7(b)}
%   \end{subfigure}
%   \includegraphics[width=1.1\linewidth]{moviecs_treebankcs_stats.png}
  \caption{\small Distribution across overall sentence lengths and distribution across lengths of continuous English spans in \moviecs and \treebankcs.}
  \label{fig:stats}
  %\vspace{-15pt}
\end{figure}

\setcounter{table}{1}
\begin{table}[b!]
%\vspace{-10pt}
\fontsize{8}{10}\selectfont
\begin{center}
    \begin{tabular}{cccc}
        \toprule
        Quantity/Metric & \moviecs & \treebankcs & \combinedcs \\
        \midrule
        $|$Train$|$ & 15509 & 5914 & 21423\\
        $|$Test$|$ & 1500 & 1000 & 2500\\
        $|$Valid$|$ & 500 & 500 & 1000\\
        \# Tokens & 196300 & 87979 & 284279\\
        \# Hindi Sentences & 9290 & 5292 & 14582 \\
        \# NEs & 4342 & 4810 & 9152 \\
        Fraction of NEs & 0.0221 & 0.0547 & 0.0322 \\
        M-Index & 0.5542 & 0.6311 & 0.5774 \\
        I-Index & 0.2852 & 0.3434 & 0.3023 \\
        \bottomrule
    \end{tabular}
    \caption{\small Key statistics of CS datasets.}
    \label{tab::metrics}
\end{center}
%\vspace{-10pt}
\end{table}
\combinedcs comprises a second subset of CS sentences, \treebankcs, that was crowdsourcing using MTurk. We extracted $5292$ monolingual Hindi sentences (with sentence lengths less than or equal to 15 words) from the publicly available Hindi Dependency Treebank that contains dependency parses.%
\footnote{\url{http://ltrc.iiit.ac.in/treebank_H2014/}}
These annotations parse each Hindi sentence into chunks, where a chunk is defined as a minimal, non recursive phrase. Turkers were asked to convert at least one Hindi chunk into English. This was done in an attempt to elicit longer spans of English segments within each sentence. Figure~\ref{fig:stats} shows the sentence length distributions for \moviecs and \treebankcs, along with histograms accumulating English segments of different lengths in both subsets. We clearly see a larger fraction of English segments with lengths within the range [2-6] in \treebankcs compared to \moviecs.
\masktodo{SK:add the linguistically correct terminology here and how exactly chunks were identified. DONE - I took the definition of a chunk from here - \url{http://verbs.colorado.edu/hindiurdu/guidelines_docs/Chunk-POS-Annotaion-Guidelines.doc}  phrase (partial structure) consisting of correlated, inseparable words/entities, such that the intra-chunk dependencies are not distorted}

%Overall, \combinedcs consists of a total of $24523$ CS sentences corresponding to $14582$ monolingual Hindi sentences. More details about the training/validation/test splits are mentioned in Table~\ref{tab::metrics}. 

Table~\ref{tab::metrics} provides detailed statistics of the new CS dataset. We also report two metrics proposed by~\citet{CS_metrics} to measure the amount of code-switching present in this new corpus. Monolingual Index (M-Index) is a value between 0 and 1 that quantifies the amount of mixing between languages (0 denotes a purely monolingual corpus and 1 denotes equal mixing from both languages) and I-Index measures the fraction of switching points in the corpus. We observe \treebankcs exhibits higher M-index and I-index values compared to \moviecs indicating more code-switching overall. \combinedcs also contains a non-trivial number of named entities (NEs) which are replaced by an NE tag in all our language modeling experiments.

\subsection{Other Datasets}
\label{section:other_datasets}

\noindent \textbf{Parallel Hindi-English Text.} As described in Section~\ref{section:experiments}, \model uses parallel text for supervised training. For this purpose, we use the IIT Bombay English-Hindi Corpus~\cite{IITB-parallel} containing parallel Hindi-English text. We also construct a larger parallel corpus using text from the OpenSubtitles (\opus) corpus~\cite{Lison2016OpenSubtitles2016EL} that is more conversational and hence more similar in style to \moviecs. We chose \textasciitilde 1 million English sentences (\opus-EN), where each sentence contained an embedded clause or a subordinate clause to support the construction of EMT lines. We used the Google Translate API to obtain Hindi translations for all these sentences (\opus-HI). Henceforth, we use \opus to refer to this parallel corpus of \opus-EN paired with \opus-HI. We extracted 318K sentences from the IITB corpus after thresholding on length (5-15) and considering overlap in vocabulary with \opus. (One could avoid the use of an external service like Google Translate and use existing parallel text~\cite{zhang2020improving}) in conjunction with a word aligner to construct EMT lines. \opus,  being more conversational in style, turns out to be a better pretraining corpus. A detailed comparison of these choices is described in Appendix~\ref{sec:appendix_add_experiments}.) \\

%{\color{red} An additional parallel En-Hi dataset (\opushundred) is used to contrast the effect of synthetic EMT lines generated using a different strategy. This has been described in detail in Appendix-\ref{sec:appendix_add_experiments}.} \\ %Our resulting total vocabulary size was $127514$.\\
%\vspace{-5pt}
\noindent \textbf{Synthetic CS Datasets.} As mentioned in Section~\ref{ssec:synthCS}, we use two simple techniques LEX and EMT to generate synthetic CS text, which in turn is used to train \model in an unsupervised training phase. For each Hindi monolingual sentence in \opus, we generate two LEX and two EMT synthetic CS sentences giving us \opus-LEX and \opus-EMT, respectively. We also generate five LEX and five EMT lines for each monolingual sentence in \combinedcs. In order to generate EMT lines, we first translate the monolingual Hindi sentences in \combinedcs to English using Google Translate and then follow the EMT generation scheme. This results in two datasets, \combinedcs-LEX and \combinedcs-EMT, which appear in later evaluations. (Appendix~\ref{sec:appendix_emt} contains more details about EMT applied to OPUS and \combinedcs.)\\

%\vspace{-5pt}
\noindent \textbf{Datasets from existing approaches.} (I) VACS~\cite{vacs} is a hierarchical variational autoencoder-based model designed to generate CS text. We train two VACS models, one on \combinedcs (VACSv1) and the other on \opus-EMT followed by \combinedcs (VACSv2). 
%The following configuration of hyperparameters is used for the training - hidden dimension is 1200, embedding size of 300, learning rate is 1.0. Both models when trained on \combinedcs converge in around 300,000 steps. VACS (v2) is trained for ~2M steps on \opus-EMT as the first step.
(II) \citet{garg-etal-2018-code} use SeqGAN~\cite{seqgan} -- a GAN-based sequence generation model -- to generate CS sentences by providing an RNNLM as the generator. As with VACS, we train two SeqGAN\footnote{\url{https://github.com/suragnair/seqGAN}} models, one on \combinedcs (SeqGANv1) and one on \opus-EMT followed by \combinedcs (SeqGANv2).
%\noindent \textbf{SeqGAN.} SeqGAN-\cite{seqgan} consists of a generator RNN and a discriminator network trained as a binary classifier to distinguish between real and generated sequences, training the generative model using policy gradients and using the discriminator to determine the reward function. \\
Samples are drawn from both SeqGAN and VACS by first drawing a random sample from the standard normal distribution in the learned latent space and then decoding via an RNN-based generator for SeqGAN and a VAE-based decoder for VACS. We sample \textasciitilde 2M lines for each dataset to match the size of the other synthetic datasets.

\section{Experiments and Results}
\label{section:experiments}

First, we investigate various training curricula to train \model and identify the best training strategy by evaluating BLEU scores on the test set of \combinedcs (\S\ref{ssec:tcsquality}). Next, we compare the output from \model with synthetic CS text generated by other methods (\S\ref{ssec:comparisons}). We approach this via language modeling (\S\ref{section:lm_modelling}), human evaluations (\S\ref{section:human_eval}) and two downstream tasks\textemdash Natural Language Inference and Sentiment Analysis\textemdash involving real CS text (\S\ref{section:GLUCoS}). Apart from these tasks, we also present four different objective evaluation metrics to evaluate synthetic CS text: BERTScore, Accuracy of a BERT-based classifier and two diversity scores (\S\ref{section:BERTScore}).  

\subsection{Improving Quality of \model Outputs}
\label{ssec:tcsquality}

Table~\ref{tab::bleuscores} shows the importance of various training curricula in training \model; these models are evaluated using BLEU~\cite{BLEU} scores computed with the ground-truth CS sentences for the test set of \combinedcs. We start with supervised pretraining of \model using the two parallel datasets we have in hand -- IITB and \opus (System $\mathbb{A}$). $\mathbb{A}$ is then further finetuned with real CS text in \combinedcs.  The improvements in BLEU scores moving from System $\mathbb{O}$ (trained only on \combinedcs) to System $\mathbb{B}$ illustrate the benefits of pretraining \model using Hindi-English parallel text. 
\begin{table}[t!]
%\vspace{-5pt}
% \small
\fontsize{9}{10}\selectfont
\setlength\tabcolsep{4pt}
% \begin{minipage}{.5\textwidth}
\begin{center}
    \begin{tabular}{llcc}
        \toprule
        & Curriculum & Training & BLEU \\
        & & & (HI \textrightarrow CS) \\
        \midrule
        $\mathbb{O}$ & \combinedcs & S & 19.18 \\
        \midrule
        $\mathbb{A}$ & IITB + \opus & S & 1.51 \\
        $\mathbb{B}$ & $\mathbb{A}$ $|$ \combinedcs & S & 27.84 \\
        \midrule \midrule
        $\mathbb{C}$ & $\mathbb{A}$ $|$ \opus-HI + \opus-LEX & U & 15.23 \\
        $\mathbb{D}$ & $\mathbb{A}$ $|$ \opus-HI + \opus-EMT & U & 17.73 \\
        \midrule \midrule
        $\mathbb{C}_1$ & $\mathbb{C}$ $|$ \combinedcs & U & 32.71 \\
        $\mathbb{C}_2$ & $\mathbb{C}$ $|$ \combinedcs & S & 39.53 \\
        \midrule \midrule
        $\mathbb{D}_1$ & $\mathbb{D}$ $|$ \combinedcs & U & 35.52 \\
        $\mathbb{D}_2$ & $\mathbb{D}$ $|$ \combinedcs & S & 43.15 \\
        \midrule \bottomrule 
        \addlinespace
    \end{tabular}
\end{center}
% \end{minipage}
%\vspace{-10pt}
\caption{\small BLEU score on (HI \textrightarrow CS) for different curricula \\ measured on \combinedcs(test). The first column gives names to each training curriculum. $\mathbb{A}$ | X represents starting with model denoted by $\mathbb{A}$ and further training using dataset(s) X. ``S" and ``U" refer to supervised and unsupervised training phases, respectively.}
\label{tab::bleuscores}
\end{table}
\begin{figure}[b!]
    %\vspace{-15pt}
    \centering
    \includegraphics[width=0.5\textwidth]{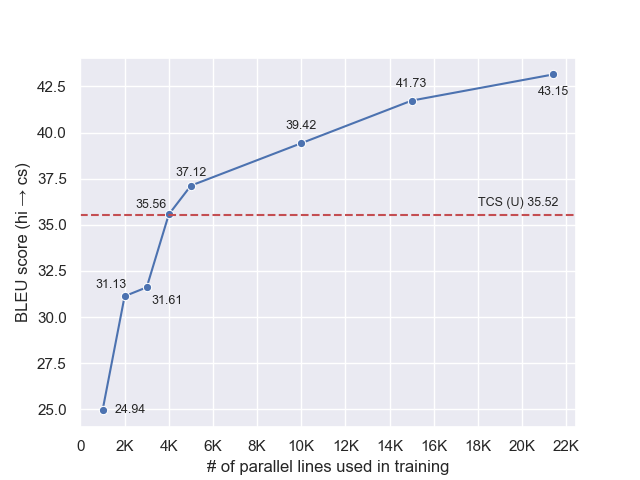}
    \caption{\small Variation of BLEU score with amount of \combinedcs parallel training data.}
    \label{fig:varybleu}
    %\vspace{-10pt}
\end{figure}

Systems $\mathbb{C}$ and $\mathbb{D}$ in Table~\ref{tab::bleuscores} use our synthetic CS datasets \opus-LEX and \opus-EMT, respectively. These systems are further finetuned on \combinedcs using both unsupervised and supervised training objectives to give $\mathbb{C}_1$, $\mathbb{C}_2$, $\mathbb{D}_1$ and $\mathbb{D}_2$, respectively. Comparing these four systems with System $\mathbb{B}$ shows the importance of using synthetic CS for pretraining. Further, comparing $\mathbb{C}_1$ against $\mathbb{D}_1$ and $\mathbb{C}_2$ against $\mathbb{D}_2$, we observe that \opus-EMT is indeed a better choice for pretraining compared to \opus-LEX. Also, supervised finetuning with \combinedcs is clearly superior to unsupervised finetuning. Henceforth, Systems $\mathbb{D}_1$ and $\mathbb{D}_2$ will be referred to as \modelunsupervised and \modelsupervised, respectively.

While having access to parallel CS data is an advantage, we argue that the benefits of having parallel data only marginally increase after a threshold. Figure~\ref{fig:varybleu} shows how BLEU scores vary when changing the amount of parallel CS text used to train $\mathbb{D}_2$. We observe that BLEU increases substantially when we increase CS data from 1000 lines to 5000 lines, after which there is a trend of diminishing returns. We also find that $\mathbb{D}_1$ (that uses the data in \combinedcs as non-parallel text) is as good as the model trained using 4000 lines of parallel text.
% We further analyse the effect of using more CS data from a different domain (as in HindiBlogCS) as non-parallel text. HindiBlogCS primarily being from technical blogs  contains a large number of words that are not in our model's vocabulary ($~$5\% OOV rate). Inspite of this, the addition of HindiBlogCS text in the final finetuning stage (D4) leads to a benefit of more than 4 BLEU points compared to D1.

\subsection{Comparing \model with Other Synthetic CS}
\label{ssec:comparisons}

\subsubsection{Language Modeling}
\label{section:lm_modelling}

We use text generated by our model to train a language model (LM) and evaluate perplexities on the test set of \combinedcs to show how closely sentences from \model mimic real CS text. We use a state-of-the-art RNNLM model AWD-LSTM-LM~\citet{awd-lstm-lm} as a blackbox LM and only experiment with different training datasets. The model uses three LSTM layers of 1200 hidden units with weight tying and 300-dimensional word embeddings. In initial runs, we trained our language model on the large parallel/synthetic CS datasets and finetuned on the \combinedcs data. However, this training strategy was prone to overfitting on \combinedcs data. To counter this problem of forgetting during the pretrain-finetuning steps, we adopted the Mix-review strategy proposed by~\citet{mix_review}. The training sentences from \combinedcs remain constant through the epochs and the amount of pretraining data is exponentially decayed with each epoch. %With an exponential decay, the pretraining batches decrease faster, due to which the model is able to iterate more through the finetuning data. 
This greatly alleviates the forgetting problem in our model, and leads to better overall perplexities. Additional details about these LMs are provided in Appendix~\ref{sec:appendix_lm_exp}.\\

%However, even after performing better during pretraining, we were not able achieve considerable improvements during finetuning. For comparison, the test perplexity on \combinedcs using OPUS+\modelsupervised was 305.07 and using OPUS+\modelunsupervised was 321.23. \masktodo{complete this. Done} To counter this problem of forgetting during pretrain-finetuning setups, we use the mix-review technique followed by ~\citet{mix_review}. The finetuning data - train sentences from \combinedcs - is kept constant through the epochs and the pretraining data is reduced in each epoch. The total number of batches in each epoch therefore keep decreasing. We experimented with keeping a linear decrease of the pretraining batches as compared to an exponential decay as the epochs progress. We found that an exponential decay greatly improved over the performance of a linear decrease. This could be because following an exponential decay, the pretraining batches decrease faster, due to which the model is able to iterate more through the finetuning data. This ensures that the forgetting problem in our model is largely alleviated, leading to better perplexities. \\
%
\begin{table}[hbt!]
%\vspace{-15pt}
\fontsize{8}{10}\selectfont
\begin{center}
    \begin{tabular}{cccc}
        \toprule
        \multirow{2}{*}{Pretraining Corpus} & \multirow{2}{*}{$\mid \text{Train}\mid$} & Test PPL & Test PPL \\
        & & \opus & \combinedcs \\
        \midrule
        % \opus & 2.03M & 53.09 & 341.10\\
        \opus + \opus-LEX & 4.00M & 56.83 & 332.66\\
        \opus + \opus-EMT & 4.03M & 55.56 & 276.56\\
        \arrayrulecolor{black!30}\midrule
        \opus + VACSv1 & 4.05M & 64.77 & 335.79\\
        \opus + VACSv2 & 4.05M & 62.41 & 321.12\\
        \opus + SeqGANv1 & 4.03M & 57.32 & 336.62\\
        \opus + SeqGANv2 & 4.03M & 56.50 & 317.81\\
        \arrayrulecolor{black!30}\midrule
        % \opus + \modellex & 4.03M & 57.24 & 268.54\\
        \opus + \modelunsupervised & 3.99M & 57.45 & 271.19\\
        % \opus + \modelsimalign & 4.03M & 60.01 & 314.28\\
        \opus + \modelsupervised & 3.96M & 56.28 & \textbf{254.37}\\
        \bottomrule
       
    \end{tabular}
        \caption{\small Test perplexities on \combinedcs using different pretraining datasets.} 
         \label{tab:lm}
\end{center}
\end{table}

%\vspace{-15pt}

Table~\ref{tab:lm} shows test perplexities using different training curricula and data generated using two prior approaches, VACS and SeqGAN. Sentences generated using \model yield the largest reductions in test perplexities, compared to all other approaches. 
%\todo{The \opus PPL numbers can be omitted from Table 4.}
%LEX, EMT, VACS (v1 and v2), SeqGAN (v1 and v2) \model sentences yields the largest reductions in test perplexities.
%The sentences generated by \modelsimalign on \opus are of slightly inferior quality because \opus wasn't part of its training curriculum. 
%Note also that the test perplexities on \opus remain fairly stable even with introducing CS text.

\subsubsection{Human Evaluation}
\label{section:human_eval}
\begin{table}[b!]
% \fontsize{10}{10}\selectfont
% \scriptsize
\small
\centering
\setlength{\belowcaptionskip}{-10pt}
\begin{tabular}{cccc}
\toprule
Method & Syntactic & Semantic & Naturalness \\
\midrule
%\multicolumn{4}{c}{Lines sampled from complete dataset} \\
%\midrule
% Real 
% & 4.36\begin{scriptsize}$\pm$0.76\end{scriptsize} 
% & 4.39\begin{scriptsize}$\pm$0.80\end{scriptsize} 
% & 4.20\begin{scriptsize}$\pm$1.00\end{scriptsize} 
% \\
% \modelsupervised 
% &  4.29\begin{scriptsize}$\pm$0.84\end{scriptsize} 
% &  4.30\begin{scriptsize}$\pm$0.89\end{scriptsize} 
% &  4.02\begin{scriptsize}$\pm$1.16\end{scriptsize}
% \\
% \modelunsupervised 
% &  3.96\begin{scriptsize}$\pm$1.06\end{scriptsize} 
% &  3.93\begin{scriptsize}$\pm$1.13\end{scriptsize} 
% &  3.52\begin{scriptsize}$\pm$1.45\end{scriptsize}
% \\
% EMT 
% &  3.47\begin{scriptsize}$\pm$1.25\end{scriptsize} 
% &  3.53\begin{scriptsize}$\pm$1.23\end{scriptsize} 
% &  2.66\begin{scriptsize}$\pm$1.49\end{scriptsize}
% \\
% LEX 
% &  3.10\begin{scriptsize}$\pm$2.16\end{scriptsize} 
% &  3.05\begin{scriptsize}$\pm$1.35\end{scriptsize} 
% &  2.01\begin{scriptsize}$\pm$1.32\end{scriptsize} 
% \\
% \midrule
% \multicolumn{4}{c}{Lines sampled from TEST + VALID} \\
% \midrule
Real 
& 4.47\begin{scriptsize}$\pm$0.73\end{scriptsize} 
& 4.47\begin{scriptsize}$\pm$0.76\end{scriptsize} 
& 4.27\begin{scriptsize}$\pm$1.06\end{scriptsize} 
\\
\modelsupervised 
&  4.21\begin{scriptsize}$\pm$0.92\end{scriptsize} 
&  4.14\begin{scriptsize}$\pm$0.99\end{scriptsize} 
&  3.77\begin{scriptsize}$\pm$1.33\end{scriptsize}
\\
\modelunsupervised 
&  4.06\begin{scriptsize}$\pm$1.06\end{scriptsize} 
&  4.01\begin{scriptsize}$\pm$1.12\end{scriptsize} 
&  3.58\begin{scriptsize}$\pm$1.46\end{scriptsize}
\\
EMT 
&  3.57\begin{scriptsize}$\pm$1.09\end{scriptsize} 
&  3.48\begin{scriptsize}$\pm$1.14\end{scriptsize} 
&  2.80\begin{scriptsize}$\pm$1.44\end{scriptsize}
\\
LEX 
&  2.91\begin{scriptsize}$\pm$1.11\end{scriptsize} 
&  2.87\begin{scriptsize}$\pm$1.19\end{scriptsize} 
&  1.89\begin{scriptsize}$\pm$1.14\end{scriptsize} 
\\
\bottomrule
\end{tabular}
\caption{\small Mean and standard deviation of scores (between 1 and 5) from 3 annotators for 150 samples from 5 datasets.}
%averaged over three human annotators. Out of 200 sentences there were 46 sentences in [1,5], 92 sentences in [6,10] and 62 sentences in [11,15] for each dataset.}
    \label{tab::human}
\end{table}

We evaluated the quality of sentences generated by \model using a human evaluation study. We sampled 150 sentences each, using both \modelunsupervised and \modelsupervised, starting from monolingual Hindi sentences in the evaluation sets of \combinedcs. The sentences were chosen such that they were consistent with the length distribution of \textsc{\combinedcs}. 
%(i.e., 14\% with a length in [1,5], 44\% sentences in [6,10] and 41\% sentences in [11,15]). 
For the sake of comparison, corresponding to the above-mentioned 150 monolingual Hindi samples, we also chose 150 CS sentences each from \combinedcs-LEX and \combinedcs-EMT. Along with the ground-truth CS sentences from \combinedcs, this resulted in a total of 750 sentences.%
\footnote{We only chose CS sentences from \model that did not exactly match the ground-truth CS text.}
These sentences were given to three linguistic experts in Hindi and they were asked to provide scores ranging between 1 and 5 (1 for worst, 5 for best) under three heads: ``Syntactic correctness", ``Semantic correctness" and ``Naturalness". 
%At least 40 out of these hundred sentences were sampled from the test and validation splits of \combinedcs, to ensure a fair comparison amongst the different methods.
Table~\ref{tab::human} shows that the sentences generated using \modelsupervised and \modelunsupervised are far superior to the EMT and LEX sentences on all three criteria. \modelsupervised is quite close in overall quality to the real sentences and \modelunsupervised fares worse, but only by a small margin. %Appendix~\ref{sec:appendix_human_eval} provides more details about an alternate human evaluation study which yields very similar results.
\setcounter{table}{6}

\begin{figure}
    \centering
    \includegraphics[width=0.48\textwidth]{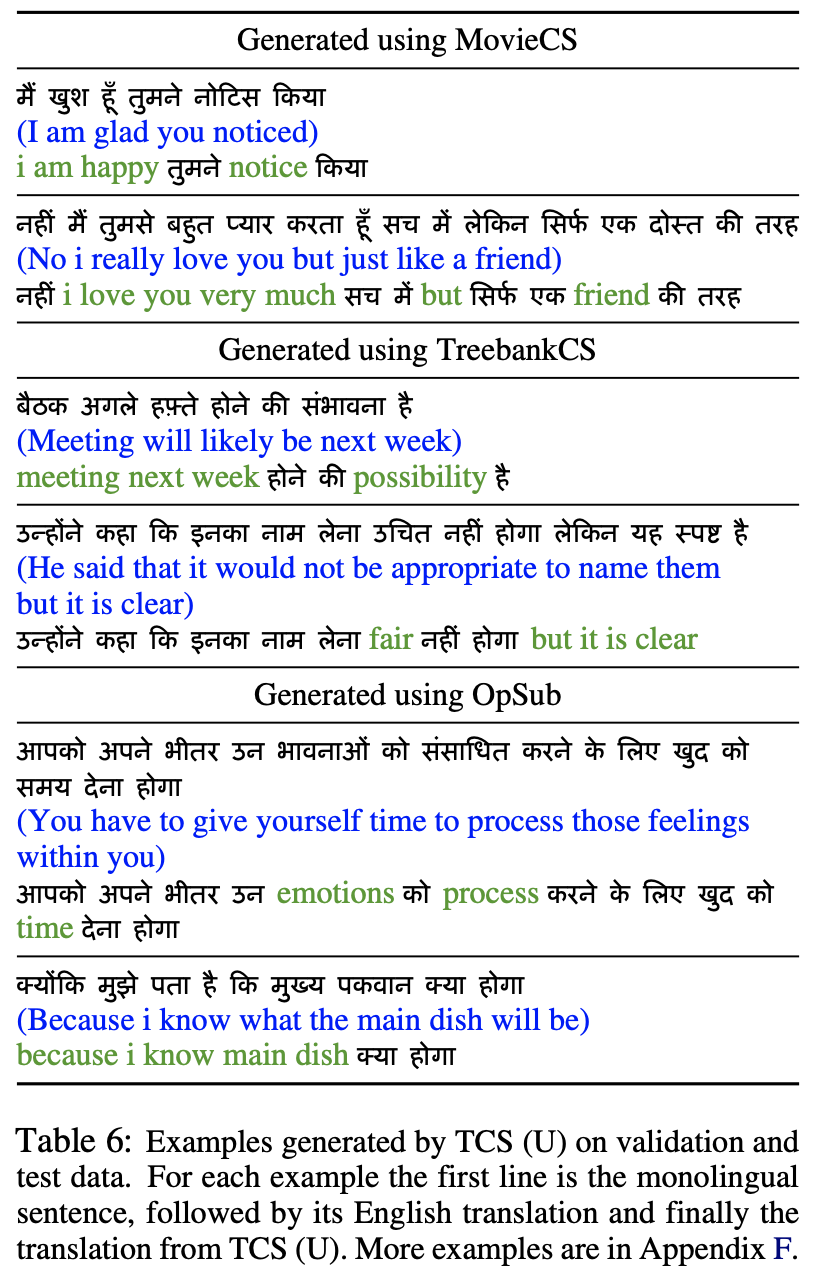}
\end{figure}

Table 6 shows some illustrative examples of code-switching using \modelunsupervised on test samples. We also show some examples of code-switching within monolingual sentences from \opus. We observe that the model is able to introduce long contiguous spans of English words (e.g. ``meeting next week", ``but it is clear", etc.). The model also displays the ability to meaningfully switch multiple times within the same sentence (e.g., ``i love you very much", ``but", ``friend"). There are also interesting cases of English segments that appear to be ungrammatical but make sense in the CS context (e.g., ``because i know main dish", etc.). 

%\vspace{-1pt}
\subsubsection{GLUECoS Benchmark}
\label{section:GLUCoS}
GLUECoS~\cite{GLUECoS} is an evaluation benchmark spanning six natural language tasks for code-switched English-Hindi and English-Spanish data. The authors  observe that M-BERT~\cite{MBERT} consistently outperforms cross-lingual embedding techniques. Furthermore, pretraining M-BERT on small amounts of code-switched text improves its performance in most cases. For our evaluation, we select two tasks that require semantic understanding: Natural Language Inference (NLI) and Sentiment Analysis (SA). %Different techniques for generating code-switched text are compared by pretraining M-BERT on the generated text and evaluating its performance.

\begin{table}[th!]
%\vspace{-10pt}
% \fontsize{10}{10}\selectfont
% \scriptsize
\small
\centering
\setlength{\belowcaptionskip}{-10pt}
\begin{tabular}{ccc}
\toprule
Pretraining Data & NLI (Accuracy) & Sentiment \\
& & Analysis (F1) \\
\midrule
Baseline
& 57.88\begin{scriptsize}$\pm$1.22\end{scriptsize} 
& 57.97\begin{scriptsize}$\pm$0.06\end{scriptsize} 
\\
\opus-HI
&  58.47\begin{scriptsize}$\pm$0.36\end{scriptsize} 
&  58.13\begin{scriptsize}$\pm$0.25\end{scriptsize} 
\\
\opus-LEX
&  58.67\begin{scriptsize}$\pm$0.94\end{scriptsize} 
&  58.40\begin{scriptsize}$\pm$0.33\end{scriptsize} 
\\
\opus-EMT
&  58.96\begin{scriptsize}$\pm$0.70\end{scriptsize} 
&  58.79\begin{scriptsize}$\pm$0.37\end{scriptsize} 
\\
\modelsupervised
&  59.57\begin{scriptsize}$\pm$0.57\end{scriptsize} 
&  \textbf{59.39\begin{scriptsize}$\pm$0.81\end{scriptsize}}
\\
\combinedcs
&  \textbf{59.74\begin{scriptsize}$\pm$0.96\end{scriptsize}} 
&  58.77\begin{scriptsize}$\pm$0.44\end{scriptsize} 
\\
\bottomrule
\end{tabular}
\caption{\small GLUECoS Evaluation: Mean and standard deviation of scores after evaluating on 5 seeds. Baseline denotes the M-BERT model without any MLM pretraining.}
    \label{tab::gluecos}
\end{table}

We sample 100K monolingual sentences from OpSub-HI and select corresponding LEX, EMT and \modelsupervised sentences. M-BERT is then trained using the masked language modelling (MLM) objective on text from all 4 systems (including OpSub-HI) for 2 epochs. We also train M-BERT on 21K sentences from \combinedcs (real CS). Finally, these pretrained models are fine-tuned on the selected GLUECoS tasks. (More details are in Appendix~\ref{sec:appendix_gluecos}.) 

Table~\ref{tab::gluecos} lists the accuracies and F1 scores using different pretraining schemes for both NLI and sentiment analysis, respectively. Plain monolingual pretraining by itself leads to performance improvements on both tasks, presumably due to domain similarity between GLUECoS (movie scripts, social media etc.) and OpSub. As mentioned in~\citet{GLUECoS}, pretraining on CS text further improves performance for both NLI and SA. Among the synthetic methods, \modelsupervised has consistently better scores than LEX and EMT. For SA, \modelsupervised even outperforms pretraining on real CS text from \combinedcs.

%\vspace{-1pt}

\begin{table*}[b]
% \vspace{-15pt}
\fontsize{8}{10}\selectfont
\centering
\setlength{\belowcaptionskip}{-10pt}
\setlength\tabcolsep{0.5pt}
\begin{tabular*}{\hsize}{@{\extracolsep{\fill}} ccccccc@{}}
\toprule
%\multicolumn{2}{c}{\multirow{2}{*}{Model}} & \multicolumn{4}{c}{Language} %\\
\multicolumn{2}{c}{Evaluation Metric} & Real & LEX & EMT & \modelsupervised & \modelunsupervised \\
\midrule
\multirow{4}{*}{BERTScore} &
All (3500) &
0.812 &
0.796 &
0.627 &
0.764 &
0.788 \\
% \midrule
&
Mono (3434) &
0.812 &
0.782 &
0.623 &
0.755 &
0.772 \\
% \midrule
&
UNK (1983) &
0.809 &
0.804 &
0.636 &
0.827 &
0.846 \\
% \midrule
&
UNK \& Mono (1857) &
\textbf{0.808} &
0.785 &
0.633 &
\textbf{0.813} &
\textbf{0.821} \\
\midrule
\multirow{2}{*}{BERT-based Classifier} &
$|$Sentences$|$ &
4767 & 
12393 &
12484 & 
12475 &
12475 \\
&
Accuracy(fake) &
\textbf{42.76} & 
96.52 &
97.83 & 
\textbf{80.31} &
\textbf{88.62} \\
\midrule
\multirow{2}{*}{Diversity} &
Gzip ($\mathbf{D}$) &
\textbf{22.13} & 
24.12 &
33.17 & 
\textbf{21.37} &
\textbf{17.59} \\
&
Self-BLEU &
\textbf{61.3} & 
29.7 &
24.6 & 
\textbf{63.6} &
\textbf{64.2} \\
\bottomrule
\end{tabular*}
\caption{\small (a) BERTScores on test split of \combinedcs. Each row corresponds to a different data filter. The numbers in parenthesis denote the number of sentences in the data after filtering. (b) Accuracies from the classifier for samples generated by various methods as being fake. The $|$Sentences$|$ refer to size of dataset for each system. \model models have the lowest accuracy among synthetic methods. (c) Diversity Scores for different techniques using Gzip and Self-BLEU based diversity measures.}
    \label{tab::evaluation_metrics}
\end{table*}

\subsection{Other Objective Evaluation Metrics}
\label{section:BERTScore}

\paragraph*{BERTScore.} BERTScore~\cite{BERTScore} is a recently-proposed evaluation metric for text generation. Similarity scores are computed between each token in the candidate sentence and each token in the reference sentence, using contextual BERT embeddings~\cite{BERToriginal} of the tokens. %Evaluation on machine translation and image captioning shows that BERTScore correlates better with human judgements. 
We use this as an additional objective metric to evaluate the quality of the sentences generated using \model. We use the real monolingual sentence as the reference and the generated CS sentence as the candidate, excluding sentences from \modelsupervised and \modelunsupervised that exactly match the real sentence. Since our data is Hindi-English CS text, we use Multilingual BERT (M-BERT)~\cite{MBERT} for high-quality multilingual representations.

%Our evaluation is performed on the test and validation sets of \combinedcs. We apply various filters to this data to clean out the noise and ensure a fair comparison.  
Table \ref{tab::evaluation_metrics} outlines our main results on the test set of \combinedcs. \model sometimes generates purely monolingual sentences. This might unfairly tilt the scores in favour of \model since the reference sentences are also monolingual. To discount for such biases, we remove sentences generated by \modelunsupervised and \modelsupervised that are purely monolingual (Row label ``Mono" in BERTScore). Sentences having \text{<UNK>} tokens (labeled ``UNK") are also filtered out since these tokens are only generated by \model for out-of-vocabulary words. ``UNK \& Mono" refers to applying both these filters.

% \model can sometimes generate purely monolingual sentences. The fraction of such sentences, however, are very small (\textasciitilde 1\%). Since BERTScore involves a similarity measure with the original monolingual sentence, such sentences might unfairly tilt the scores in favour of \model. To discount for such biases, we remove sentences generated by \modelunsupervised and \modelsupervised that are purely monolingual. This is shown in the second row in Table \ref{tab::evaluation_metrics}.

% \model generates an \text{<UNK>} token for out-of-vocabulary words. The other methods, however, involve translations and do not use \text{<UNK>} tags. Hence, we consider filtering out sentences that have one or more \text{<UNK>} tokens (in row 3); in row 4, we exclude both monolingual sentences and \text{<UNK>}s.

EMT lines consistently show the worst performance, which is primarily due to the somewhat poor quality of translations involved in generating these lines (refer to Appendix~\ref{sec:appendix_emt}). With removing both monolingual and \text{<UNK>} tokens, we observe that  \modelunsupervised and \modelsupervised yield the highest BERTScores, even outperforming the BERTScore on real data obtained from the Turkers.

% \begin{table*}
% % \vspace{-15pt}
% \fontsize{8}{10}\selectfont
% \centering
% \setlength{\belowcaptionskip}{-10pt}
% \setlength\tabcolsep{0.5pt}
% \begin{tabular*}{\hsize}{@{\extracolsep{\fill}} lccccccc@{}}
% \toprule
% %\multicolumn{2}{c}{\multirow{2}{*}{Model}} & \multicolumn{4}{c}{Language} %\\
% & Real & LEX & EMT & \modelsupervised & \modelunsupervised & \modellex & \modelsimalign \\
% \midrule
% \multirow{1}{*}{All (3500)} &
% 0.812 &
% 0.796 &
% 0.627 &
% 0.764 &
% 0.788 &
% 0.773 &
% 0.768 \\
% \midrule
% \multirow{1}{*}{$\; -$Mono (3434)} &
% 0.812 &
% 0.782 &
% 0.623 &
% 0.755 &
% 0.772 &
% 0.769 &
% 0.753 \\
% \midrule
% \multirow{1}{*}{$\; -$UNK (1983)} &
% 0.809 &
% 0.804 &
% 0.636 &
% 0.827 &
% 0.846 &
% 0.832 &
% 0.829 \\
% \midrule
% \multirow{1}{*}{$\; -$UNK \& Mono (1857)} &
% \textbf{0.808} &
% \textbf{0.785} &
% \textbf{0.633} &
% \textbf{0.813} &
% \textbf{0.821} &
% \textbf{0.817} &
% \textbf{0.822} \\
% \bottomrule
% \end{tabular*}
% \caption{\small BERTScores on test and validation data of \combinedcs with different data filters. Each row corresponds to a filter applied to the data. The numbers in parenthesis denote the number of sentences in the data after filtering.}
%     \label{tab::bertscores}
% \end{table*}

\paragraph*{BERT-based Classifier.}  In this evaluation, we use M-BERT~\cite{MBERT} to build a classifier that distinguishes real CS sentences from synthetically generated ones (fake). When subject to examples from high-quality generators, the classifier should find it hard to tell apart real from fake samples. We add a fully connected layer over the M-BERT base architecture that takes the [CLS] token as its input to predict the probability of the sentence being real or fake. Fake sentences are drawn from the union of \modelunsupervised, \modelsupervised, \combinedcs-LEX and \combinedcs-EMT. In order to alleviate the class imbalance problem, we oversample the real sentences by a factor of 5 and shuffle the data. The model converges after training for 5 epochs. We see in Table~\ref{tab::evaluation_metrics} that the classification accuracy of whether a sample is fake or not is lowest for the outputs from \model among the different generation techniques. 
% (i.e. 5 sentences generated using each scheme on monolingual sentences of \combinedcs test and valid split)
%We see that the classifier assigns higher probability to sentences generated from \modelunsupervised (0.1059) and \modelsupervised (0.1464) as compared to LEX (0.0227) and EMT (0.0118). 

% \label{seche accurification ofasentences generetcy of cltio n:assBERT_classifier}
% \begin{table}[t!]
% \small
% \begin{center}
%     \begin{tabular}{ccc}
%         \toprule
%         Method & $|$Test$|$ &  Accuracy (fake)\\
%         \midrule
%         \combinedcs & 4767 & 42.76 \\
%         \midrule
%         LEX & 12393 & 96.52 \\
%         EMT & 12484  & 97.83 \\
%         \modelunsupervised & 12475  & 88.62 \\
%         \modelsupervised & 12475  & 80.31 \\
%         \modellex &  12475 & 84.17 \\
%         \modelsimalign &  12475 & 82.98 \\
%         \bottomrule
%     \end{tabular}
%     \caption{\small Accuracies from the classifier for samples generated by various methods as being fake. \modelunsupervised and \modelsupervised have the lowest accuracy among synthetic methods.}
%     \label{tab::classifier}
% \end{center}
% \end{table}

\paragraph*{Measuring Diversity.}
\label{section:diversity}

%For each monolingual sentence in \combinedcs, multiple CS variants can be generated as shown in Table~\ref{tab::example}. 
We are interested in finding out how diverse the predictions from \model are. We propose a simple measure of diversity in the CS variants that is based on how effectively sentences can be compressed using the gzip utility.%
\footnote{\url{http://www.gzip.org/}}
%
%gzip uses a combination of Huffman coding and an adaptive codebook based on the Lempel-Ziv algorithm~\cite{LZencoding}. 
We considered using Byte Pair Encoding (BPE)~\cite{bpe} as a measure of data compression. However, BPE operates at the level of individual words. Two word sequences ``w1 w2 w3" and ``w3 w2 w1" would be identically compressed by a BPE tokenizer. We would ideally like to account for such diversity and not discard this information. gzip uses Lempel-Ziv coding~\cite{LZencoding} that considers substrings of characters during compression, thus allowing for diversity in word ordering to be captured.

Our diversity measure $\mathbf{D}$ is simply the following: For a given set of CS sentences, run gzip on each sentence individually and sum the resulting file sizes ($S_1$). Next, paste all the CS sentences into a single file and run gzip on it to get a file of size $S_2$. Then, $\mathbf{D} = S_1 - S_2$. Smaller $\mathbf{D}$ scores indicate larger diversity. If the variants of a sentence are dissimilar to one another and hence very diverse, then $S_2$ would be large thus leading to smaller values of $\mathbf{D}$. Table~\ref{tab::evaluation_metrics} shows the diversity scores for different techniques. Both \modelsupervised and \modelunsupervised have a higher diversity score compared to LEX and EMT. \modelunsupervised exceeds even the responses received via MTurk (Real) in diversity. We note here that diversity, by itself, is not necessarily a desirable trait. Our goal is to generate sentences that are diverse while being natural and semantically meaningful. The latter properties for text from \modelsupervised and \modelunsupervised have already been verified in our human evaluation study.

%{\color{red} Generating CS from monolingual requires both retaining the meaning and generating appropriate switch points. 
\newcite{Texygen} propose self-BLEU score as a metric to evaluate the diversity of generated data. However, using self-BLEU is slightly problematic in our setting as systems like LEX that switch words at random positions would result in low self-BLEU (indicating high diversity). This is indeed the case, as shown in Table~\ref{tab::evaluation_metrics} - LEX, EMT give lower self-BLEU scores as compared to \model. However, note that the scores of the \model models are comparable to that of real CS data.

\section{Conclusions}
\label{sec:conclusion}
In this work, we present a neural translation model for CS text that transduces monolingual Hindi sentences into realistic Hindi-English CS text. Text generated by our model is evaluated using a number of different objective metrics, along with LM, NLI and sentiment analysis tasks, and a detailed human evaluation study. The role of synthetic data in training such models merits a more detailed investigation which we leave for future work. 

\section{Acknowledgements}
We thank all the anonymous reviewers for their constructive feedback which helped improve the presentation of this work. We also thank all the volunteers who helped with the collection of CS text that is released as part of our dataset, \combinedcs.

\bibliographystyle{acl_natbib}

\begin{thebibliography}{51}
\expandafter\ifx\csname natexlab\endcsname\relax\def\natexlab#1{#1}\fi

\bibitem[{Adel et~al.(2015)Adel, Vu, Kirchhoff, Telaar, and
  Schultz}]{adel2015syntactic}
Heike Adel, Ngoc~Thang Vu, Katrin Kirchhoff, Dominic Telaar, and Tanja Schultz.
  2015.
\newblock Syntactic and semantic features for code-switching factored language
  models.
\newblock \emph{IEEE/ACM transactions on audio, speech, and language
  Processing}, 23(3):431--440.

\bibitem[{Adel et~al.(2013)Adel, Vu, Kraus, Schlippe, Li, and
  Schultz}]{adel2013recurrent}
Heike Adel, Ngoc~Thang Vu, Franziska Kraus, Tim Schlippe, Haizhou Li, and Tanja
  Schultz. 2013.
\newblock Recurrent neural network language modeling for code switching
  conversational speech.
\newblock In \emph{2013 IEEE International Conference on Acoustics, Speech and
  Signal Processing}, pages 8411--8415. IEEE.

\bibitem[{Amazon(2005)}]{mturk}
Amazon. 2005.
\newblock \href {http://www.mturk.com/} {Amazon mechanical turk website}.
\newblock Visited on 2020-01-02.

\bibitem[{Bhat et~al.(2016)Bhat, Choudhury, and Bali}]{Bhat:16}
Gayatri Bhat, Monojit Choudhury, and Kalika Bali. 2016.
\newblock Grammatical constraints on intra-sentential code-switching:from
  theories to working models.
\newblock \emph{arXiv:1612.04538}.

\bibitem[{Chang et~al.(2019)Chang, Chuang, and Lee}]{chang-etal}
Ching-Ting Chang, Shun-Po Chuang, and Hung-Yi Lee. 2019.
\newblock \href {https://doi.org/10.21437/Interspeech.2019-3214}
  {{Code-Switching Sentence Generation by Generative Adversarial Networks and
  its Application to Data Augmentation}}.
\newblock In \emph{Proc. Interspeech 2019}, pages 554--558.

\bibitem[{Conneau et~al.(2017)Conneau, Lample, Ranzato, Denoyer, and
  J{\'e}gou}]{conneau2017word}
Alexis Conneau, Guillaume Lample, Marc'Aurelio Ranzato, Ludovic Denoyer, and
  Herv{\'e} J{\'e}gou. 2017.
\newblock Word translation without parallel data.
\newblock \emph{arXiv preprint arXiv:1710.04087}.

\bibitem[{Devlin et~al.(2018)Devlin, Chang, Lee, and Toutanova}]{BERToriginal}
Jacob Devlin, Ming{-}Wei Chang, Kenton Lee, and Kristina Toutanova. 2018.
\newblock \href {http://arxiv.org/abs/1810.04805} {{BERT:} pre-training of deep
  bidirectional transformers for language understanding}.
\newblock \emph{CoRR}, abs/1810.04805.

\bibitem[{Dhar et~al.(2018)Dhar, Kumar, and
  Shrivastava}]{dhar-etal-2018-enabling}
Mrinal Dhar, Vaibhav Kumar, and Manish Shrivastava. 2018.
\newblock \href {https://www.aclweb.org/anthology/W18-3817} {Enabling
  code-mixed translation: Parallel corpus creation and {MT} augmentation
  approach}.
\newblock In \emph{Proceedings of the First Workshop on Linguistic Resources
  for Natural Language Processing}, pages 131--140, Santa Fe, New Mexico, USA.
  Association for Computational Linguistics.

\bibitem[{Gage(1994)}]{bpe}
Philip Gage. 1994.
\newblock A new algorithm for data compression.
\newblock \emph{C Users J.}, 12(2):23–38.

\bibitem[{Gao et~al.(2019)Gao, Feng, Liu, Hou, Pan, and Ma}]{gao2019code}
Yingying Gao, Junlan Feng, Ying Liu, Leijing Hou, Xin Pan, and Yong Ma. 2019.
\newblock Code-switching sentence generation by bert and generative adversarial
  networks.
\newblock In \emph{INTERSPEECH}, pages 3525--3529.

\bibitem[{Garg et~al.(2018{\natexlab{a}})Garg, Parekh, and
  Jyothi}]{garg-etal-2018-code}
Saurabh Garg, Tanmay Parekh, and Preethi Jyothi. 2018{\natexlab{a}}.
\newblock \href {https://doi.org/10.18653/v1/D18-1346} {Code-switched language
  models using dual {RNN}s and same-source pretraining}.
\newblock In \emph{Proceedings of the 2018 Conference on Empirical Methods in
  Natural Language Processing}, pages 3078--3083, Brussels, Belgium.
  Association for Computational Linguistics.

\bibitem[{Garg et~al.(2018{\natexlab{b}})Garg, Parekh, and
  Jyothi}]{garg2018dual}
Saurabh Garg, Tanmay Parekh, and Preethi Jyothi. 2018{\natexlab{b}}.
\newblock Dual language models for code switched speech recognition.
\newblock \emph{Proceedings of Interspeech}, pages 2598--2602.

\bibitem[{Gautam et~al.(2021)Gautam, Kodali, Gupta, Goel, Shrivastava, and
  Kumaraguru}]{gautam2021comet}
Devansh Gautam, Prashant Kodali, Kshitij Gupta, Anmol Goel, Manish Shrivastava,
  and Ponnurangam Kumaraguru. 2021.
\newblock Comet: Towards code-mixed translation using parallel monolingual
  sentences.
\newblock In \emph{Proceedings of the Fifth Workshop on Computational
  Approaches to Linguistic Code-Switching}, pages 47--55.

\bibitem[{Gupta et~al.(2021)Gupta, Vavre, and Sarawagi}]{gupta2021training}
Abhirut Gupta, Aditya Vavre, and Sunita Sarawagi. 2021.
\newblock Training data augmentation for code-mixed translation.
\newblock In \emph{Proceedings of the 2021 Conference of the North American
  Chapter of the Association for Computational Linguistics: Human Language
  Technologies}, pages 5760--5766.

\bibitem[{Guzm{\'a}n et~al.(2017)Guzm{\'a}n, Ricard, Serigos, Bullock, and
  Toribio}]{CS_metrics}
Gualberto~A. Guzm{\'a}n, Joseph Ricard, Jacqueline Serigos, Barbara~E. Bullock,
  and Almeida~Jacqueline Toribio. 2017.
\newblock Metrics for modeling code-switching across corpora.
\newblock In \emph{INTERSPEECH}.

\bibitem[{Hara et~al.(2018)Hara, Adams, Milland, Savage, Callison-Burch, and
  Bigham}]{MTurkWage}
Kotaro Hara, Abigail Adams, Kristy Milland, Saiph Savage, Chris Callison-Burch,
  and Jeffrey~P. Bigham. 2018.
\newblock \href {https://doi.org/10.1145/3173574.3174023} {\emph{A Data-Driven
  Analysis of Workers' Earnings on Amazon Mechanical Turk}}, page 1–14.
  Association for Computing Machinery, New York, NY, USA.

\bibitem[{He et~al.(2021)He, Liu, Cho, Ott, Liu, Glass, and Peng}]{mix_review}
Tianxing He, Jun Liu, Kyunghyun Cho, Myle Ott, Bing Liu, James Glass, and
  Fuchun Peng. 2021.
\newblock \href {https://www.aclweb.org/anthology/2021.eacl-main.95} {Analyzing
  the forgetting problem in pretrain-finetuning of open-domain dialogue
  response models}.
\newblock In \emph{Proceedings of the 16th Conference of the European Chapter
  of the Association for Computational Linguistics: Main Volume}, pages
  1121--1133, Online. Association for Computational Linguistics.

\bibitem[{Jalili~Sabet et~al.(2020)Jalili~Sabet, Dufter, Yvon, and
  Sch{\"u}tze}]{sabet2020simalign}
Masoud Jalili~Sabet, Philipp Dufter, Fran{\c{c}}ois Yvon, and Hinrich
  Sch{\"u}tze. 2020.
\newblock \href {https://doi.org/10.18653/v1/2020.findings-emnlp.147}
  {{S}im{A}lign: High quality word alignments without parallel training data
  using static and contextualized embeddings}.
\newblock In \emph{Findings of the Association for Computational Linguistics:
  EMNLP 2020}, pages 1627--1643, Online. Association for Computational
  Linguistics.

\bibitem[{Khanuja et~al.(2020)Khanuja, Dandapat, Srinivasan, Sitaram, and
  Choudhury}]{GLUECoS}
Simran Khanuja, Sandipan Dandapat, Anirudh Srinivasan, Sunayana Sitaram, and
  Monojit Choudhury. 2020.
\newblock \href {https://doi.org/10.18653/v1/2020.acl-main.329} {{GLUEC}o{S}:
  An evaluation benchmark for code-switched {NLP}}.
\newblock In \emph{Proceedings of the 58th Annual Meeting of the Association
  for Computational Linguistics}, pages 3575--3585, Online. Association for
  Computational Linguistics.

\bibitem[{Kunchukuttan et~al.(2017)Kunchukuttan, Mehta, and
  Bhattacharyya}]{IITB-parallel}
Anoop Kunchukuttan, Pratik Mehta, and Pushpak Bhattacharyya. 2017.
\newblock \href {http://arxiv.org/abs/1710.02855} {The {IIT} bombay
  english-hindi parallel corpus}.
\newblock \emph{CoRR}, abs/1710.02855.

\bibitem[{Lample et~al.(2018{\natexlab{a}})Lample, Conneau, Denoyer, and
  Ranzato}]{unmt}
Guillaume Lample, Alexis Conneau, Ludovic Denoyer, and Marc'Aurelio Ranzato.
  2018{\natexlab{a}}.
\newblock \href {https://openreview.net/forum?id=rkYTTf-AZ} {Unsupervised
  machine translation using monolingual corpora only}.
\newblock In \emph{International Conference on Learning Representations}.

\bibitem[{Lample et~al.(2018{\natexlab{b}})Lample, Ott, Conneau, Denoyer, and
  Ranzato}]{lample-etal-2018-phrase}
Guillaume Lample, Myle Ott, Alexis Conneau, Ludovic Denoyer, and Marc{'}Aurelio
  Ranzato. 2018{\natexlab{b}}.
\newblock \href {https://doi.org/10.18653/v1/D18-1549} {Phrase-based {\&}
  neural unsupervised machine translation}.
\newblock In \emph{Proceedings of the 2018 Conference on Empirical Methods in
  Natural Language Processing}, pages 5039--5049, Brussels, Belgium.
  Association for Computational Linguistics.

\bibitem[{Lee and Li(2020)}]{lee-li-2020-modeling}
Grandee Lee and Haizhou Li. 2020.
\newblock \href {https://doi.org/10.18653/v1/2020.acl-main.80} {Modeling
  code-switch languages using bilingual parallel corpus}.
\newblock In \emph{Proceedings of the 58th Annual Meeting of the Association
  for Computational Linguistics}, pages 860--870, Online. Association for
  Computational Linguistics.

\bibitem[{Li and Fung(2013)}]{li2013improved}
Ying Li and Pascale Fung. 2013.
\newblock Improved mixed language speech recognition using asymmetric acoustic
  model and language model with code-switch inversion constraints.
\newblock In \emph{2013 IEEE International Conference on Acoustics, Speech and
  Signal Processing}, pages 7368--7372. IEEE.

\bibitem[{Li and Fung(2014{\natexlab{a}})}]{Li2014CodeSL}
Ying Li and Pascale Fung. 2014{\natexlab{a}}.
\newblock Code switch language modeling with functional head constraint.
\newblock \emph{2014 IEEE International Conference on Acoustics, Speech and
  Signal Processing (ICASSP)}, pages 4913--4917.

\bibitem[{Li and Fung(2014{\natexlab{b}})}]{li-fung-2014-language}
Ying Li and Pascale Fung. 2014{\natexlab{b}}.
\newblock \href {https://doi.org/10.3115/v1/D14-1098} {Language modeling with
  functional head constraint for code switching speech recognition}.
\newblock In \emph{Proceedings of the 2014 Conference on Empirical Methods in
  Natural Language Processing ({EMNLP})}, pages 907--916, Doha, Qatar.
  Association for Computational Linguistics.

\bibitem[{Lison and Tiedemann(2016)}]{Lison2016OpenSubtitles2016EL}
Pierre Lison and J{\"o}rg Tiedemann. 2016.
\newblock Opensubtitles2016: Extracting large parallel corpora from movie and
  tv subtitles.
\newblock In \emph{LREC}.

\bibitem[{Liu et~al.(2020)Liu, Gu, Goyal, Li, Edunov, Ghazvininejad, Lewis, and
  Zettlemoyer}]{liu2020multilingual}
Yinhan Liu, Jiatao Gu, Naman Goyal, Xian Li, Sergey Edunov, Marjan
  Ghazvininejad, Mike Lewis, and Luke Zettlemoyer. 2020.
\newblock Multilingual denoising pre-training for neural machine translation.
\newblock \emph{Transactions of the Association for Computational Linguistics},
  8:726--742.

\bibitem[{Merity et~al.(2018)Merity, Keskar, and Socher}]{awd-lstm-lm}
Stephen Merity, Nitish~Shirish Keskar, and Richard Socher. 2018.
\newblock \href {https://openreview.net/forum?id=SyyGPP0TZ} {Regularizing and
  optimizing {LSTM} language models}.
\newblock In \emph{International Conference on Learning Representations}.

\bibitem[{Mikolov et~al.(2013)Mikolov, Sutskever, Chen, Corrado, and
  Dean}]{embeddings}
Tomas Mikolov, Ilya Sutskever, Kai Chen, Greg Corrado, and Jeffrey Dean. 2013.
\newblock \href {http://arxiv.org/abs/1310.4546} {Distributed representations
  of words and phrases and their compositionality}.
\newblock \emph{CoRR}, abs/1310.4546.

\bibitem[{Papineni et~al.(2002)Papineni, Roukos, Ward, and Zhu}]{BLEU}
Kishore Papineni, Salim Roukos, Todd Ward, and Wei-Jing Zhu. 2002.
\newblock \href {https://doi.org/10.3115/1073083.1073135} {Bleu: A method for
  automatic evaluation of machine translation}.
\newblock In \emph{Proceedings of the 40th Annual Meeting on Association for
  Computational Linguistics}, ACL ’02, page 311–318, USA. Association for
  Computational Linguistics.

\bibitem[{Pires et~al.(2019)Pires, Schlinger, and Garrette}]{MBERT}
Telmo Pires, Eva Schlinger, and Dan Garrette. 2019.
\newblock \href {https://doi.org/10.18653/v1/P19-1493} {How multilingual is
  multilingual {BERT}?}
\newblock In \emph{Proceedings of the 57th Annual Meeting of the Association
  for Computational Linguistics}, pages 4996--5001, Florence, Italy.
  Association for Computational Linguistics.

\bibitem[{Poplack(1979)}]{Poplack1979SometimesIS}
Shana Poplack. 1979.
\newblock “sometimes i'll start a sentence in spanish y termino en
  espa{\~n}ol”: Toward a typology of code-switching.

\bibitem[{Pratapa et~al.(2018)Pratapa, Bhat, Choudhury, Sitaram, Dandapat, and
  Bali}]{pratapa2018language}
Adithya Pratapa, Gayatri Bhat, Monojit Choudhury, Sunayana Sitaram, Sandipan
  Dandapat, and Kalika Bali. 2018.
\newblock \href
  {https://www.microsoft.com/en-us/research/publication/language-modeling-code-mixing-role-linguistic-theory-based-synthetic-data/}
  {Language modeling for code-mixing: The role of linguistic theory based
  synthetic data}.
\newblock In \emph{Proceedings of ACL 2018}. ACL.

\bibitem[{Rizvi et~al.(2021)Rizvi, Srinivasan, Ganu, Choudhury, and
  Sitaram}]{Rizvi2021GCMAT}
Mohd Sanad~Zaki Rizvi, Anirudh Srinivasan, T.~Ganu, M.~Choudhury, and Sunayana
  Sitaram. 2021.
\newblock Gcm: A toolkit for generating synthetic code-mixed text.
\newblock In \emph{EACL}.

\bibitem[{Samanta et~al.(2019)Samanta, Reddy, Jagirdar, Ganguly, and
  Chakrabarti}]{vacs}
Bidisha Samanta, Sharmila Reddy, Hussain Jagirdar, Niloy Ganguly, and Soumen
  Chakrabarti. 2019.
\newblock \href {https://doi.org/10.24963/ijcai.2019/719} {A deep generative
  model for code switched text}.
\newblock In \emph{Proceedings of the Twenty-Eighth International Joint
  Conference on Artificial Intelligence, {IJCAI-19}}, pages 5175--5181.
  International Joint Conferences on Artificial Intelligence Organization.

\bibitem[{Sankoff(1998)}]{sankoff_1998}
David Sankoff. 1998.
\newblock \href {https://doi.org/10.1017/S136672899800011X} {A formal
  production-based explanation of the facts of code-switching}.
\newblock \emph{Bilingualism: Language and Cognition}, 1(1):39–50.

\bibitem[{Sennrich et~al.(2015)Sennrich, Haddow, and Birch}]{sennrich}
Rico Sennrich, Barry Haddow, and Alexandra Birch. 2015.
\newblock \href {http://arxiv.org/abs/1511.06709} {Improving neural machine
  translation models with monolingual data}.
\newblock \emph{CoRR}, abs/1511.06709.

\bibitem[{Solorio et~al.(2021)Solorio, Chen, Black, Diab, Sitaram, Soto, and
  Yilmaz}]{calcs-2021-approaches}
Thamar Solorio, Shuguang Chen, Alan~W. Black, Mona Diab, Sunayana Sitaram,
  Victor Soto, and Emre Yilmaz, editors. 2021.
\newblock \href {https://www.aclweb.org/anthology/2021.calcs-1.0}
  {\emph{Proceedings of the Fifth Workshop on Computational Approaches to
  Linguistic Code-Switching}}. Association for Computational Linguistics,
  Online.

\bibitem[{Tiedemann(2012)}]{tiedemann-2012-parallel}
J{\"o}rg Tiedemann. 2012.
\newblock \href
  {http://www.lrec-conf.org/proceedings/lrec2012/pdf/463_Paper.pdf} {Parallel
  data, tools and interfaces in {OPUS}}.
\newblock In \emph{Proceedings of the Eighth International Conference on
  Language Resources and Evaluation ({LREC}'12)}, pages 2214--2218, Istanbul,
  Turkey. European Language Resources Association (ELRA).

\bibitem[{Vaswani et~al.(2017)Vaswani, Shazeer, Parmar, Uszkoreit, Jones,
  Gomez, Kaiser, and Polosukhin}]{transformer}
Ashish Vaswani, Noam Shazeer, Niki Parmar, Jakob Uszkoreit, Llion Jones,
  Aidan~N. Gomez, Lukasz Kaiser, and Illia Polosukhin. 2017.
\newblock \href {http://arxiv.org/abs/1706.03762} {Attention is all you need}.
\newblock \emph{CoRR}, abs/1706.03762.

\bibitem[{Vinyals et~al.(2015)Vinyals, Fortunato, and
  Jaitly}]{vinyals2015pointer}
Oriol Vinyals, Meire Fortunato, and Navdeep Jaitly. 2015.
\newblock Pointer networks.
\newblock In \emph{Advances in neural information processing systems}, pages
  2692--2700.

\bibitem[{Vu et~al.(2012)Vu, Lyu, Weiner, Telaar, Schlippe, Blaicher, Chng,
  Schultz, and Li}]{vu2012first}
Ngoc~Thang Vu, Dau-Cheng Lyu, Jochen Weiner, Dominic Telaar, Tim Schlippe,
  Fabian Blaicher, Eng-Siong Chng, Tanja Schultz, and Haizhou Li. 2012.
\newblock A first speech recognition system for mandarin-english code-switch
  conversational speech.
\newblock In \emph{2012 IEEE International Conference on Acoustics, Speech and
  Signal Processing (ICASSP)}, pages 4889--4892. IEEE.

\bibitem[{Winata et~al.(2018)Winata, Madotto, Wu, and
  Fung}]{learn-to-code-switch}
Genta~Indra Winata, Andrea Madotto, Chien{-}Sheng Wu, and Pascale Fung. 2018.
\newblock \href {http://arxiv.org/abs/1810.10254} {Learn to code-switch: Data
  augmentation using copy mechanism on language modeling}.
\newblock \emph{CoRR}, abs/1810.10254.

\bibitem[{Winata et~al.(2019)Winata, Madotto, Wu, and Fung}]{pointer-net}
Genta~Indra Winata, Andrea Madotto, Chien-Sheng Wu, and Pascale Fung. 2019.
\newblock Code-switched language models using neural based synthetic data from
  parallel sentences.
\newblock In \emph{CoNLL}.

\bibitem[{Yeh et~al.()Yeh, Huang, Sun, and Lee}]{yeh2010integrated}
Ching~Feng Yeh, Chao~Yu Huang, Liang~Che Sun, and Lin~Shan Lee.
\newblock An integrated framework for transcribing mandarin-english code-mixed
  lectures with improved acoustic and language modeling.
\newblock In \emph{2010 7th International Symposium on Chinese Spoken Language
  Processing}, pages 214--219. IEEE.

\bibitem[{Yu et~al.(2017)Yu, Zhang, Wang, and Yu}]{seqgan}
Lantao Yu, Weinan Zhang, Jun Wang, and Yong Yu. 2017.
\newblock Seqgan: Sequence generative adversarial nets with policy gradient.
\newblock In \emph{Thirty-First AAAI Conference on Artificial Intelligence}.

\bibitem[{Zhang et~al.(2020)Zhang, Williams, Titov, and
  Sennrich}]{zhang2020improving}
Biao Zhang, Philip Williams, Ivan Titov, and Rico Sennrich. 2020.
\newblock \href {https://doi.org/10.18653/v1/2020.acl-main.148} {Improving
  massively multilingual neural machine translation and zero-shot translation}.
\newblock In \emph{Proceedings of the 58th Annual Meeting of the Association
  for Computational Linguistics}, pages 1628--1639, Online. Association for
  Computational Linguistics.

\bibitem[{Zhang* et~al.(2020)Zhang*, Kishore*, Wu*, Weinberger, and
  Artzi}]{BERTScore}
Tianyi Zhang*, Varsha Kishore*, Felix Wu*, Kilian~Q. Weinberger, and Yoav
  Artzi. 2020.
\newblock \href {https://openreview.net/forum?id=SkeHuCVFDr} {Bertscore:
  Evaluating text generation with bert}.
\newblock In \emph{International Conference on Learning Representations}.

\bibitem[{Zhu et~al.(2018)Zhu, Lu, Zheng, Guo, Zhang, Wang, and Yu}]{Texygen}
Yaoming Zhu, Sidi Lu, Lei Zheng, Jiaxian Guo, Weinan Zhang, Jun Wang, and Yong
  Yu. 2018.
\newblock \href {https://doi.org/10.1145/3209978.3210080} {Texygen: A
  benchmarking platform for text generation models}.
\newblock In \emph{The 41st International ACM SIGIR Conference on Research \&
  Development in Information Retrieval}, SIGIR '18, page 1097–1100, New York,
  NY, USA. Association for Computing Machinery.

\bibitem[{{Ziv} and {Lempel}(1977)}]{LZencoding}
J.~{Ziv} and A.~{Lempel}. 1977.
\newblock A universal algorithm for sequential data compression.
\newblock \emph{IEEE Transactions on Information Theory}, 23(3):337--343.

\end{thebibliography}

\clearpage

\appendix

\section{MTurk Task Details}
\label{sec:appendix_mturk}
\begin{figure}[H]
  \centering
  \fbox{\includegraphics[width=0.9\linewidth]{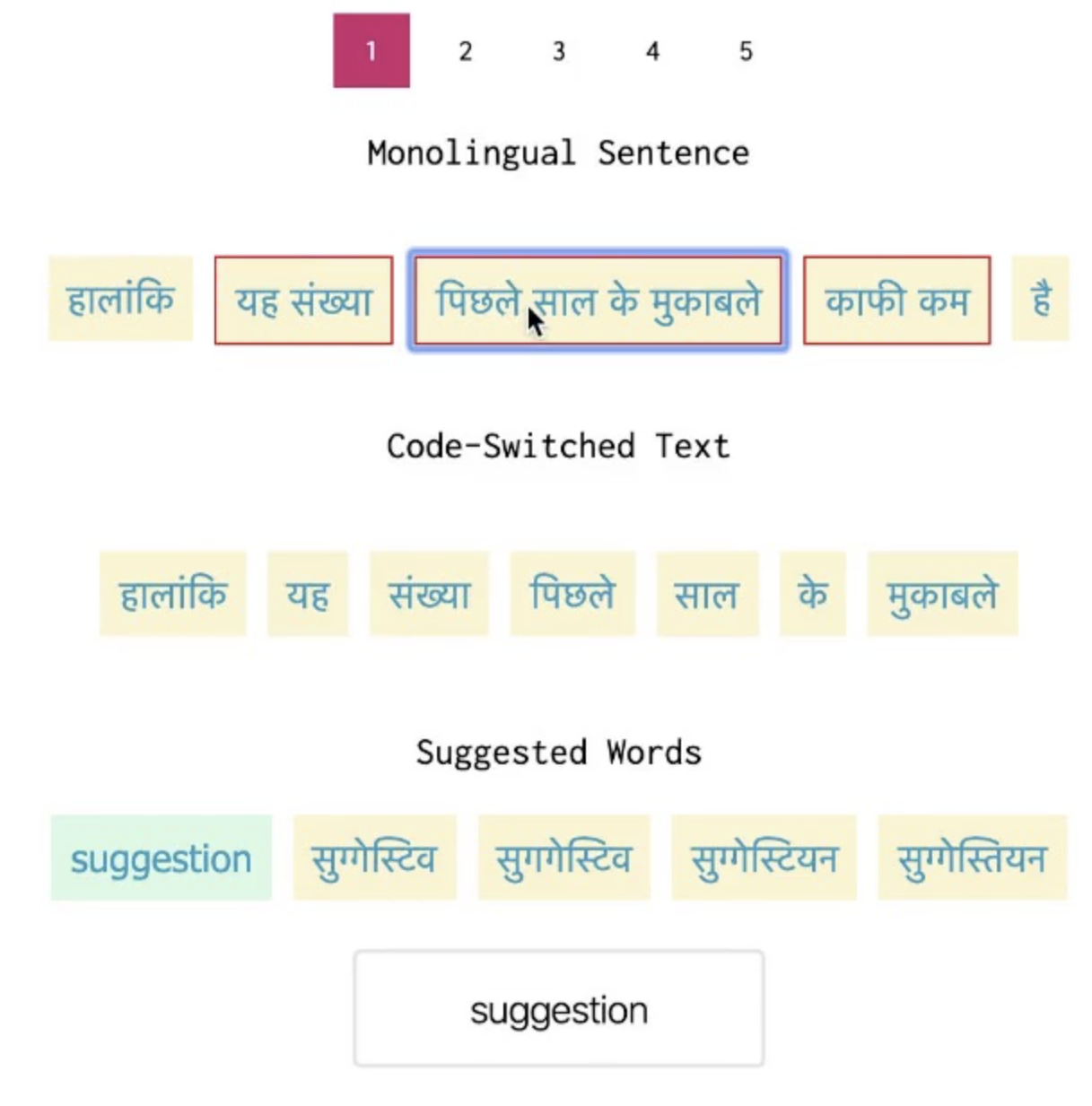}}
  \caption{\small A snapshot of the web interface used to collect \moviecs and \treebankcs data via Amazon Mechanical Turk.}
  \label{fig:mturk_portal}
\end{figure}

Figure~\ref{fig:mturk_portal} depicts the portal used to collect data using Amazon's Mechanical Turk platform. The collection was done in two rounds, first for \moviecs and then for \treebankcs. With \treebankcs, the sentences were first divided into chunks and the Turkers were provided with a sentence grouped into chunks as shown in Figure~\ref{fig:mturk_portal}. They were required to switch at least one chunk in the sentence entirely to English so as to ensure a longer span of English words in the resulting CS sentence. A suggestion box converted transliterated Hindi words into Devanagari  and also provided English suggestions to aid the workers in completing their task. With \moviecs, since there were no chunk labels associated with the sentences, they were tokenized into words. 

On MTurk, we selected workers with HIT approval rate of 90\% and location restricted to countries with significant Hindi speakers - Australia, Bahrain, Canada, India, Kuwait, Malaysia, Mauritius, Myanmar, Nepal, Netherlands, New Zealand, Oman, Pakistan, Qatar, Saudi Arabia, Singapore, South Africa, Sri Lanka, Thailand, United Arab Emirates, United Kingdom, United States of America. It was clearly specified in the guidelines that the task must be attempted by native Hindi speakers. Each response was manually checked before approving. Turkers were paid \$0.15 for working on 5 sentences (roughly takes 3-4 minutes). This amounts to \$2.25-\$3/hr which is in the ballpark of a median hourly wage on MTurk of \textasciitilde\$2/hr \cite{MTurkWage}.

\section{EMT lines generation}
\label{sec:appendix_emt}
Following the methodology described in~\cite{Bhat:16}, we apply clause substitution methodology to produce EMT sentences. To create \opus-EMT, we start with the gold English sentence that contains either embedded sentence clauses (S) or subordinate clauses (SBAR) and swap one or more of them with their Hindi translations to produce an EMT synthetic CS sentence. Due to the lack of gold English translations available for \combinedcs sentences, we used the Google Translate API to first acquire their English translation. Many of the sentences in \combinedcs are shorter in length and do not contain the abovementioned clauses. So, we also considered inverted declarative sentence clauses (SINV), inverted question clauses (SQ) and direct question clauses (SBARQ) in addition to S and SBAR. In case none of the clause level tags were present, we considered the following phrase level tags as switching candidates:  Noun Phrase (NP), Verb Phrase (VP), Adjective Phrase (ADJP) and Adverb Phase (ADVP). Owing to the shorter length and lack of clause-level tags, we switch only one tag per sentence for \combinedcs-EMT. The choice of which clause to switch was made empirically by observing  what switches caused the resulting sentence to resemble a naturally occurring CS sentence. One can also use the toolkit provided by \citet{Rizvi2021GCMAT} for generating EMT lines.

\masktodo{Add process of generation of EMT lines for \combinedcs. Done}

\iffalse
\section{Comparison to \citet{chang-etal}}
\label{sec:chang}
We list down three major limitations of the model proposed by \citet{chang-etal}. The authors use the SEAME corpus to train their GAN-based model, which is an order of magnitude larger (106K sentences) in size compared to our dataset. This is antithetical to our main premise that we may not have access to a lot of CS text during training. Also for each token in the monolingual sentence, their  model outputs a 0 or 1 denoting whether to keep the token as-is or translate it, respectively. This methodology forces the generated text to be exactly the same length as the input monolingual text unlike our approach. And finally token-level translations which are not context-aware do not typically give rise to realistic CS text; TCS does not do token-level translations and uses the entire sequence as in neural translation models.
\fi

\section{Implementation Details: TCS}
\label{sec:appendix_tcs}

As an initialisation step, we learn the token embeddings~\cite{embeddings} on the same corpus using skipgram. The embedding dimension was set to be 256 and the encoder-decoder layers share these lookup tables. Adam optimiser with a learning rate of $0.0001$ was used to train the model. Validation BLEU scores on (HI $\rightarrow$ ENG/CS) translations and (EN $\rightarrow$ HI $\rightarrow$ EN) reconstructions were used as metrics to save the best model for \modelsupervised and \modelunsupervised, respectively.

\section {Human Evaluation}
\label{sec:appendix_human_eval}
The 150 samples evaluated in Table~\ref{tab::human} were taken entirely from test/validation splits. We undertook an alternate human evaluation experiment involving 100 real CS sentences and its corresponding CS sentences using LEX, EMT, \modelunsupervised and \modelsupervised. Out of these 100 sentences, 40 of them came entirely from the test and validation splits and the remaining 60 are training sentences which we filtered to make sure that sentences generated by \modelsupervised and \modelunsupervised never exactly matched the real CS sentence.  The table below (Table \ref{tab::human2}) reports the evaluations on the complete set of 100 sentences from 5 datasets. We observe that the trend remains exactly the same as in Table~\ref{tab::human}, with \modelsupervised being very close to real CS sentences in its evaluation and \modelunsupervised trailing behind \modelsupervised. 

\begin{table}[H]
\small
\centering
\setlength{\belowcaptionskip}{-10pt}
\begin{tabular}{cccc}
\toprule
Method & Syntactic & Semantic & Naturalness \\
\midrule
% \multicolumn{4}{c}{Lines sampled from complete dataset} \\
\midrule
Real 
& 4.36\begin{scriptsize}$\pm$0.76\end{scriptsize} 
& 4.39\begin{scriptsize}$\pm$0.80\end{scriptsize} 
& 4.20\begin{scriptsize}$\pm$1.00\end{scriptsize} 
\\
\modelsupervised 
&  4.29\begin{scriptsize}$\pm$0.84\end{scriptsize} 
&  4.30\begin{scriptsize}$\pm$0.89\end{scriptsize} 
&  4.02\begin{scriptsize}$\pm$1.16\end{scriptsize}
\\
\modelunsupervised 
&  3.96\begin{scriptsize}$\pm$1.06\end{scriptsize} 
&  3.93\begin{scriptsize}$\pm$1.13\end{scriptsize} 
&  3.52\begin{scriptsize}$\pm$1.45\end{scriptsize}
\\
EMT 
&  3.47\begin{scriptsize}$\pm$1.25\end{scriptsize} 
&  3.53\begin{scriptsize}$\pm$1.23\end{scriptsize} 
&  2.66\begin{scriptsize}$\pm$1.49\end{scriptsize}
\\
LEX 
&  3.10\begin{scriptsize}$\pm$2.16\end{scriptsize} 
&  3.05\begin{scriptsize}$\pm$1.35\end{scriptsize} 
&  2.01\begin{scriptsize}$\pm$1.32\end{scriptsize} 
\\

\bottomrule
\end{tabular}
\caption{\small Mean and standard deviation of scores (between 1 and 5) from 3 annotators for 100 samples from 5 datasets.}
%averaged over three human annotators. Out of 200 sentences there were 46 sentences in [1,5], 92 sentences in [6,10] and 62 sentences in [11,15] for each dataset.}
    \label{tab::human2}
\end{table}

\section{Language Model Training}
\label{sec:appendix_lm_exp}
The AWD-LSTM language model was trained for 100 epochs with a batch size of 80 and a sequence length of 70 in each batch. The learning rate was set at 30. The model uses NT-ASGD, a variant of the averaged stochastic gradient method, to update the weights. The mix-review decay parameter was set to 0.9. This implies that the fraction of pretraining batches being considered at the end of $n$ epochs is $0.9^{n}$, starting from all batches initially. Two decay coefficients \{0.8, 0.9\} were tested and 0.9 was chosen based on validation perplexities.

\section{Code-switching examples}
\label{sec:appendix_cs_examples}

\begin{figure}[t!]
    \centering
    \includegraphics[width=0.48\textwidth]{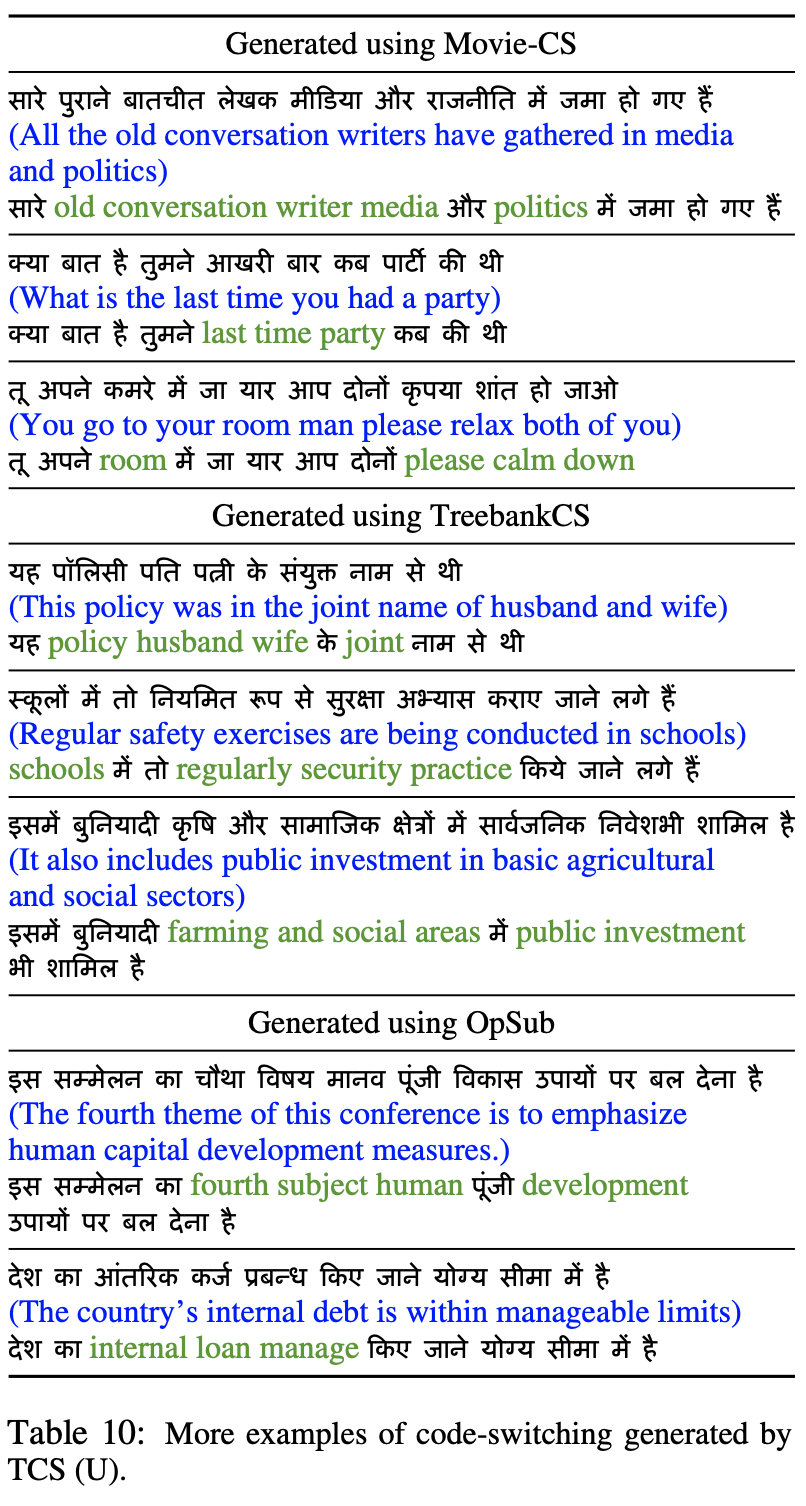}
\end{figure}
\setcounter{table}{10}

The sentences in Table 10 have been generated on the test and validation splits of \combinedcs as well as the \opus dataset. Overall, they depict how the model is able to retain context over long sentences (e.g. ``and social sectors") and perform meaningful switching over large spans of words (e.g. ``old conversation writer media", ``regularly security practices"). We also note that at times, the model uses words which are different from the natural English translations of the sentence, which are appropriate within the context of a CS sentence (e.g. the use of ``manage" instead of ``manageable").

\section{Details of GLUECoS Experiments}
\label{sec:appendix_gluecos}
For masked language modeling (MLM), we select the default parameters for the learning rate (5e-5), batch masking probability (0.15), sequence length (512). The models are trained for 2 epochs with a batch size of 4 and gradient accumulation step of 10. For task specific fine tuning we rely on the official training scripts provided by GLUECoS repository.
\footnote{https://github.com/microsoft/GLUECoS}
We train the models for 5 seed (0,1,2,3 and 4) and report mean and standard deviations of Accuracy and F1 for NLI and Sentiment Analysis respectively

\section{Additional Dataset and Experiments}
\label{sec:appendix_add_experiments}

\noindent \textbf{Dataset} The additional corpus on which experiments were performed is \opushundred \cite{zhang2020improving} which was sampled from the original OPUS corpus \cite{tiedemann-2012-parallel}. The primary difference between \opus and \opushundred is that \opus does not have manual Hindi translations of its sentences and requires the use of an external API such as Google Translate for translation. However, \opushundred has manually annotated sentences as part of the corpus. The source of \opushundred ranges from movie subtitles to GNOME documentation to the Bible. %Performance against performance using \opus described in Section \ref{section:other_datasets}.  
We extract 340K sentences from \opushundred corpus after thresholding on length (5-15). We offer this comparison of systems trained on \opus and \opushundred to show how our models fare when using two datasets that are very different in their composition. \\
%Sentences from OPUS are more conversational in nature, while the text in the IITB corpus is more formal. Thus, we used OPUS to determine our vocabulary and extracted 318K sentences from the IITB corpus that overlapped in vocabulary.

%Another Hi-En parallel corpus form multiple sources of total size of ~1.56M. (//TODO cite) We sample ~318K lines from this based on overlap in vocabulary to MONO and length thresholding. 

%Using the OpenSubtitles corpus comprising of english subtitles we choose ~1M sentences containing an embedded clause or subordinate clause and call it OPUS\_EN. We obtain the translations of these sentences using the Google Translate API to form the monolingual Hindi dataset OPUS\_HI. OPUS thus is a parallel corpus of OPUS\_EN and OPUS\_HI. \\

\noindent \textbf{LEX lines generation.} Generation of LEX lines is straightforward and requires only a bilingual lexicon. For each monolingual Hindi sentence we generate \textasciitilde5 sentences on \opushundred resulting in \opushundred-LEX (to roughly match the size of \opus-LEX). \\

\noindent \textbf{EMT lines generation.} For generation of EMT lines we have two strategies depending on the availability of tools (parsers, translation service, aligners, etc). The first strategy requires a translation service (either in-house or publicly available). We substitute the embedded clause from parse trees of English sentences with their Hindi translations. This strategy does not  require a parallel Hindi corpus and has been previously used for generating \opus-EMT and \combinedcs-EMT (Described in detail in Appendix~\ref{sec:appendix_emt}). \\

\noindent The second strategy, that is used to generate \opushundred-EMT,  requires a parallel corpus, a constituent parser in English and a word aligner between parallel sentences. \opushundred sentences are aligned using SimAlign \cite{sabet2020simalign} and embedded clauses from parse trees of English sentences are replaced by Hindi clauses using word aligners. Here again, for each monolingual Hindi sentenece we generate \textasciitilde5 EMT sentences (strategy-2) on \opushundred resulting in \opushundred-EMT.\\

\begin{table}[t!]
% \small
\fontsize{9}{10}\selectfont
\setlength\tabcolsep{4pt}
% \begin{minipage}{.5\textwidth}
\begin{center}
    \begin{tabular}{llcc}
        \toprule
        & Curriculum & X=\opus & X=\opushundred\\
        \midrule
        $\mathbb{O}$ & \combinedcs (S) &  19.18 & 19.14\\
        \midrule
        $\mathbb{A}$ & IITB + X (S) & 1.51 & 0.29\\
        $\mathbb{B}$ & $\mathbb{A}$ $|$ \combinedcs (S) &  27.84 & 25.63\\
        \midrule
        $\mathbb{C}$ & $\mathbb{A}$ $|$ X-HI + X-LEX (U) &  15.23 & 14.17\\
        $\mathbb{C}_1$ & $\mathbb{C}$ $|$ \combinedcs (U) & 32.71 & 31.48\\
        $\mathbb{C}_2$ & $\mathbb{C}$ $|$ \combinedcs (S) &  39.53 & 37.51\\
        \midrule
        $\mathbb{D}$ & $\mathbb{A}$ $|$ X-HI + X-EMT (U) & 17.73 & 15.03\\
        $\mathbb{D}_1$ & $\mathbb{D}$ $|$ \combinedcs (U) & 35.52 & 33.91\\
        $\mathbb{D}_2$ & $\mathbb{D}$ $|$ \combinedcs (S) & 43.15 & 40.32\\
        \bottomrule
        \addlinespace
    \end{tabular}
\end{center}
%\vspace{-15pt}
% \end{minipage}
\caption{\small BLEU score on (HI $\textrightarrow$ CS) for different curricula \\ measured on \combinedcs (test). $\mathbb{X}$ | Y represents starting with  model X and further training using dataset Y. Values from Table \ref{tab::bleuscores} are replicated here for ease of comparison.}
    \label{tab::bleuscores_appendix}
\end{table}
 
 \noindent \textbf{Curriculum Training Experiments.} Table~\ref{tab::bleuscores_appendix} provides a walkthrough of systems using various training curricula that are evaluated for two different choices of datasets - \opus vs \opushundred differing in the generation of EMT lines. The models are evaluated using BLEU~\cite{BLEU} scores computed on the test set of \combinedcs. The vocabulary is generated by combining train sets of all datasets to be used in the curricula. It is 126,576 when X = \opus and 164,350 when X = \opushundred (\opus shows a higher overlap in vocabulary with \combinedcs compared to \opushundred).  The marginal difference in System $\mathbb{O}$ for \opus and \opushundred is attributed to differences in the size of the vocabulary. \opus being conversational in nature, is a better pretraining corpus compared to \opushundred as seen from System $\mathbb{A}$, the sources of the latter being GNOME documentations and The Bible, apart from movie subtitles.

The results for $\mathbb{C}_1$, $\mathbb{C}_2$, $\mathbb{D}_1$, $\mathbb{D}_2$ are consistently better when X = \opus versus when X = \opushundred. We choose to highlight four models from Table~\ref{tab::bleuscores_appendix} which together demonstrate multiple use-cases of \model in Table \ref{tab:model_use_case}. \modellex refers to ($\mathbb{C}_2$, X=\opus), \modelunsupervised refers to ($\mathbb{D}_1$, X=\opus), \modelsupervised refers to ($\mathbb{D}_2$, X=\opus) and \modelsimalign refers to ($\mathbb{D}_2$, X=\opushundred).  \\

\begin{table}[b!]
\fontsize{8}{10}\selectfont
\begin{center}
    \begin{tabular}{l|l}
        \toprule
        \model Model & Use-Case \\
        \midrule
        \multirow{2}{*}{\modellex} &  Easy generation of sentences, \\ 
        & only requires a bilingual lexicon \\
        \midrule
        \multirow{2}{*}{\modelunsupervised  and \modelsupervised }
        & Requires parser and translation service \\
        & Does not require parallel data\\ \midrule
        \multirow{2}{*}{\modelsimalign} & Requires parser along with parallel data\\ 
        & Alignment can be generated \\ &using SimAlign\\
        \bottomrule
    \end{tabular}
        \caption{\small Use cases for different \model models.} 
         \label{tab:model_use_case}
\end{center}
%\vspace{-10pt}
\end{table}

\begin{table}[h]
\fontsize{8}{10}\selectfont
\begin{center}
    \begin{tabular}{cccc}
        \toprule
        \multirow{2}{*}{Pretraining Corpus} & \multirow{2}{*}{$\mid \text{Train}\mid$} & Test PPL & Test PPL \\
        & & \opus & \combinedcs \\
        \midrule
        \opus + \modellex & 4.03M & 57.24 & 268.54\\
        \opus + \modelunsupervised & 3.99M & 57.45 & 271.19\\
        \opus + \modelsimalign & 4.03M & 60.01 & 314.28\\
        \opus + \modelsupervised & 3.96M & 56.28 & \textbf{254.37}\\
        \bottomrule
       
    \end{tabular}
        \caption{\small Test perplexities on \opus and \combinedcs using different pretraining datasets.} 
         \label{tab::lm_appendix}
\end{center}
\vspace{-10pt}
\end{table}
\noindent \textbf{Language Modelling Experiments.} Table~\ref{tab::lm_appendix} shows results from LM experiments (using the same setup as in Section \ref{section:lm_modelling}). The values for \modelsupervised and \modelunsupervised have been reproduced here for ease of comparison. (Note that \modelsimalign does not perform as well as the other models since the sentences for training the language model are generated on \opus for all the models here, but \modelsimalign has been trained on \opushundred.) \\

\noindent \textbf{Evaluation Metrics.} Table~\ref{tab::evaluation_metrics_additional} shows the results of the three  objective evaluation metrics on the additional \model models. In comparison with the results in Table \ref{tab::evaluation_metrics}, we observe that \modellex and \modelsimalign perform comparably to \modelsupervised and \modelunsupervised on all metrics.

\begin{table}[H]
\fontsize{8}{10}\selectfont
\begin{center}
\setlength{\belowcaptionskip}{-10pt}
\setlength\tabcolsep{0.5pt}
\begin{tabular}{cccc}
\toprule
%\multicolumn{2}{c}{\multirow{2}{*}{Model}} & \multicolumn{4}{c}{Language} %\\
\multicolumn{2}{c}{Evaluation Metric} & \modellex & \modelsimalign \\
\midrule
\multirow{4}{*}{BERTScore} &
All (3500) &
0.773 &
0.768 \\
% \midrule
&
Mono (3434) &
0.769 &
0.753 \\
% \midrule
&
UNK (1983) &
0.832 &
0.829 \\
% \midrule
&
UNK \& Mono (1857) &
\textbf{0.817} &
\textbf{0.822} \\
\midrule
BERT-based &
$|$Sentences$|$ &
12475 &
12475 \\
Classifier &
Accuracy(fake) &
\textbf{84.17} &
\textbf{82.98} \\
\midrule

\multirow{2}{*}{Diversity} &
Gzip ($\mathbf{D}$) &
\textbf{19.62} &
\textbf{19.83} \\
&
Self-BLEU &
\textbf{56.3} &
\textbf{59.8} \\

\bottomrule
\end{tabular}
\caption{\small Evaluation metrics for the additional \model models. Please see Table~\ref{tab::evaluation_metrics} for a comparison with other models.}
    \label{tab::evaluation_metrics_additional}
\end{center}
\end{table}

\end{document}